\documentclass[final,1p,times]{elsarticle}

\usepackage{amssymb}
\usepackage{amsthm}
\usepackage[T5]{fontenc}
\usepackage[utf8]{inputenc}
\usepackage{booktabs}
\usepackage{float}
\usepackage{adjustbox}
\usepackage{caption}
\usepackage{subcaption}
\usepackage{hyperref}
\usepackage{soul}
\usepackage{soul}
\usepackage{xcolor}
\usepackage{graphicx}
\usepackage{tabularx}
\usepackage{multirow}
\usepackage{amsmath}
\usepackage{amsfonts}
\usepackage{algorithm}
\usepackage{algpseudocode}
\usepackage{url}
\usepackage{pifont}%
\usepackage{placeins}
\usepackage[title]{appendix}%

\usepackage{xcolor, color, soul}
\sethlcolor{yellow}
\usepackage[flushleft]{threeparttable}

\newcommand{\cmark}{\ding{51}}
\newcommand{\xmark}{\ding{55}}

\DeclareUnicodeCharacter{2713}{\cmark}  
\DeclareUnicodeCharacter{2717}{\xmark}  

\journal{Expert Systems with Applications}

\begin{document}

\begin{frontmatter}



\title{ViTextVQA: A Large-Scale Visual Question Answering Dataset and a Novel Multimodal Feature Fusion Method for Vietnamese Text Comprehension in Images}

\author[1,2]{Quan Van Nguyen}
\ead{21521333@gm.uit.edu.vn}

\author[1,2]{Dan Quang Tran}
\ead{21521917@gm.uit.edu.vn}

\author[1,2]{Huy Quang Pham}
\ead{21522163@gm.uit.edu.vn}

\author[1,2]{Thang Kien-Bao Nguyen}
\ead{21521432@gm.uit.edu.vn}

\author[1,2]{Nghia Hieu Nguyen}
\ead{nghiangh@uit.edu.vn}

\author[1,2]{Kiet Van Nguyen\corref{cor1}}
\ead{kietnv@uit.edu.vn}

\author[1,2]{Ngan Luu-Thuy Nguyen}
\ead{ngannlt@uit.edu.vn}

\affiliation[1]{organization={Faculty of Information Science and Engineering, University of Information Technology},
            city={Ho Chi Minh city},
            country={Vietnam}}
\affiliation[2]{organization={Vietnam National University},
            city={Ho Chi Minh city},
            country={Vietnam}}
            
\cortext[cor1]{Corresponding author}


\begin{abstract}
Visual Question Answering (VQA) is a challenging task that requires the joint understanding of natural language and visual content. While early research primarily focused on recognizing objects and scene context, it often overlooked scene text-an essential source of explicit semantic information. This paper introduces \textbf{ViTextVQA} (\textbf{Vi}etnamese \textbf{Text}-based \textbf{V}isual \textbf{Q}uestion \textbf{A}nswering), the first large-scale Vietnamese dataset specializing in text-based VQA. The dataset contains \textbf{over 16,000} images and \textbf{over 50,000} question-answer pairs. To tackle this task efficiently, \textbf{ViTextBLIP-2} (Vietnamese Text-based Bootstrapped Language-Image Model via Fine-tuning) is proposed, a novel multimodal feature fusion method designed to optimize Vietnamese text-based VQA. Experiments with state-of-the-art models highlight the importance of token ordering in OCR text for answer generation, leading to significant performance improvements. The ViTextVQA dataset is publicly available for research purposes\footnote{\url{https://huggingface.co/datasets/minhquan6203/ViTextVQA}}.
\end{abstract}


\end{frontmatter}


\section{Introduction}\label{sec1} 
In recent years, Visual Question Answering (VQA) has garnered significant attention from researchers in Computer Vision (CV) and Natural Language Processing (NLP). The appeal of VQA has surged with the advent of powerful models like Flamingo \cite{alayrac2022flamingo}, GPT-4 \cite{achiam2023gpt}, and Gemini \cite{team2023gemini}, which integrate advanced image-based question-answering capabilities. This has propelled VQA into a phase of robust global development, particularly in resource-rich languages such as English and Chinese. Numerous VQA datasets, including \cite{biten2019scene, singh2019towards, wang2022chiqa, qi2022dureadervis}, have been released, accompanied by innovative methods showcasing remarkable performance, driven largely by neural and transformer architectures \cite{biten2022latr, geigle2023mblip, kil2023prestu}, which extend beyond traditional approaches.

In Vietnamese, \citet{tran2021vivqa} introduced ViVQA, the first dataset tailored for this task, comprising 15,000 samples derived semi-automatically from VQA v2. Despite its pioneering role, ViVQA faced scrutiny for its limitations, as later detailed by \citet{nguyen2023openvivqa}. Addressing these issues, the OpenViVQA dataset emerged in 2023, designed for open-ended Vietnamese VQA with 11,199 images and 37,914 manually crafted question-answer (QA) pairs. This dataset marked a shift toward open-ended questions and answers, opening new research avenues. Subsequently, the ViOCRVQA dataset \cite{pham2025viocrvqa} scaled up efforts with nearly 30,000 images and over 120,000 QA pairs, emphasizing text-based question answering in images.

These datasets still exhibit limitations. The ViVQA dataset did not mention scene text. The OpenViVQA dataset incorporates it in about half of its samples, the free-form answers complicate objective performance assessment. The ViOCRVQA fully focuses on text-based VQA but without scene text.

\begin{figure}[htp]
\centering
\includegraphics[width=0.75\textwidth]{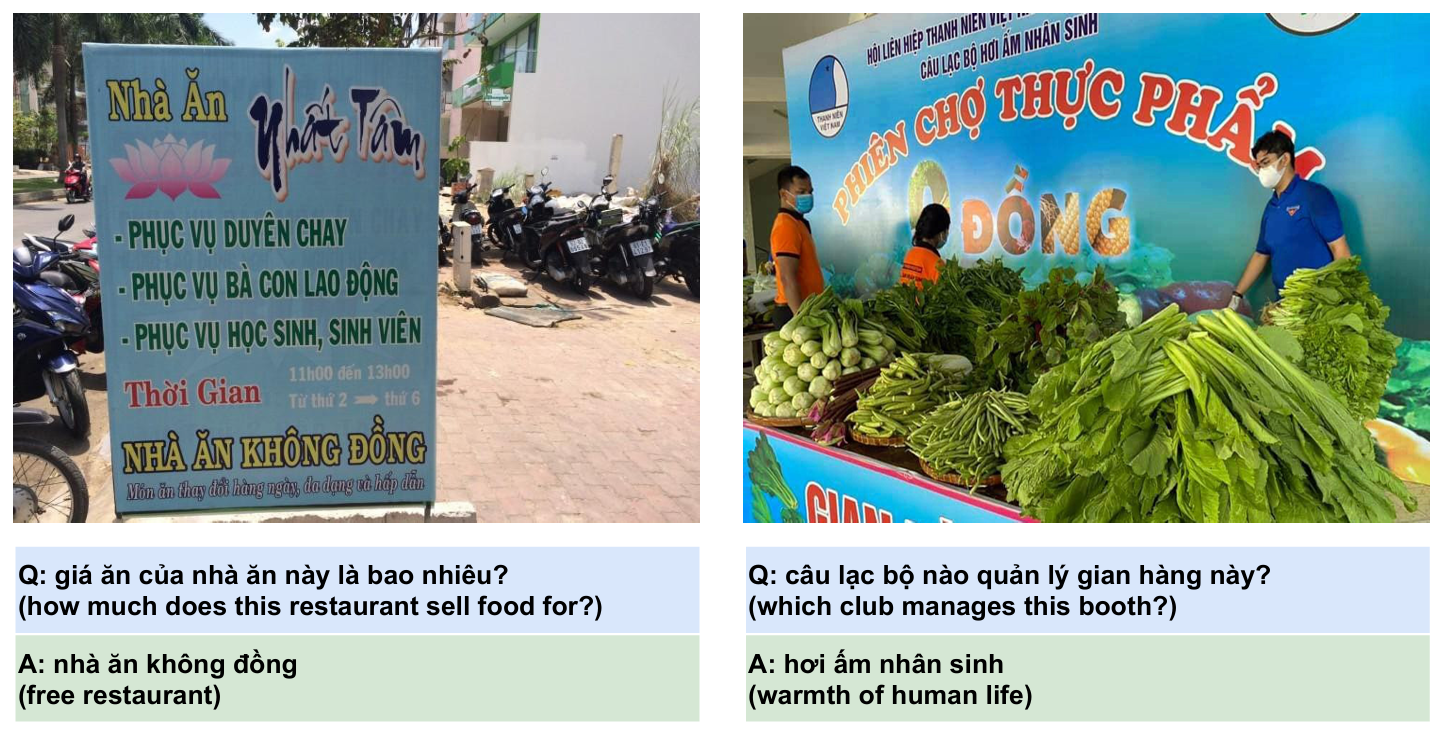}
\caption{Samples extracted from the ViTextVQA dataset.}
\label{example_data_1}
\end{figure}

To minimize these shortcomings, we introduce the \textbf{Vi}etnamese \textbf{Text}-based \textbf{V}isual \textbf{Q}uestion \textbf{A}nswering (ViTextVQA) dataset, featuring 16,762 images and 50,342 QA pairs. ViTextVQA prioritizes extracting information from scene text to tackle challenges traditional VQA models face (see Figure \ref{example_data_1}). Unlike ViVQA, which relied on semi-automatic question generation, the ViTextVQA dataset emphasizes manually curated and text-centric questions to ensure higher quality. In contrast to OpenViVQA, which included scene text in only about half of its samples, ViTextVQA provides comprehensive coverage of text-based questions. Compared with ViOCRVQA, which focuses on OCR-based VQA but omits scene text, the dataset integrates both elements, offering a more holistic benchmark. To further advance this task, we propose ViTextBLIP-2 (Vietnamese Text-based Bootstrapped Language-Image Model via Fine-tuning), a BLIP-2-based architecture optimized for Vietnamese text-based VQA. In addition to pre-trained vision and language models with a trainable Q-Former for multimodal feature fusion, ViTextBLIP-2 incorporates OCR outputs and image captions to provide both fine-grained textual details and global contextual cues, as detailed in Section \ref{sec4}.

The main contributions of this study are summarized as follows:

\begin{itemize}
    \item \textbf{A new benchmark for Vietnamese text-based VQA.} We present \textit{ViTextVQA}, a high-quality and large-scale dataset focusing on scene text understanding in Vietnamese. It contains over 16,000 images and 50,000 question-answer pairs with diverse question types and long-form responses, establishing the first comprehensive resource for this task.
    
    \item \textbf{A unified and efficient multimodal framework.} We propose \textit{ViTextBLIP-2}, a novel model that leverages frozen pre-trained components and a trainable Q-Former (initialized from BLIP-2) to enable effective integration of visual, textual, and OCR features. This design substantially improves the efficiency and adaptability of VQA systems in low-resource settings.
    
    \item \textbf{Comprehensive analysis of text layout and OCR ordering.} Through systematic experiments on ViTextVQA, we demonstrate that proper OCR text ordering (from top-left to bottom-right) significantly enhances model performance. This finding provides new insights into the importance of text layout adaptation for multilingual and low-resource VQA tasks.
\end{itemize}

This article is structured as follows: Section \ref{sec2} reviews related studies. Section \ref{sec3} details the ViTextVQA creation process and analysis. Section \ref{sec4} introduces the proposed ViTextBLIP-2. Section \ref{sec5} presents baseline models and outlines experimental design and results. Section \ref{sec6} delves into result analysis. Section \ref{sec7} explores Vietnamese-specific traits. Section \ref{sec8} presents the study’s limitations and directions for future research. Finally, Section \ref{sec9} concludes the study.

\section{Related work}\label{sec2}
\subsection{Visual Question Answering datasets}
\subsubsection{Former Visual Question Answering Datasets}

\begin{table*}[!ht]
\centering
\setlength{\tabcolsep}{10pt}
\caption{Overview of the former VQA datasets in English.}
\label{en_dataset}
\begin{adjustbox}{width=1\textwidth}
\begin{tabular}{lcccr}
\hline
\textbf{Dataset} & \textbf{Type} & \textbf{Image Source}            & \textbf{Annotation Method} & \textbf{Language} \\ \hline
DAQUAR \cite{malinowski2014multi}           & Non text      & NYU-DepthV2                      & Human annotation           & English           \\
VQA v1 \cite{antol2015vqa}           & Non text      & MS COCO                          & Human annotation           & English           \\
VQA v2 \cite{goyal2017making}           & Non text      & MS COCO                          & Human annotation           & English           \\
TextVQA \cite{singh2019towards}          & Text      & MS COCO                          & Human annotation           & English           \\
OCR-VQA \cite{mishra2019ocr}         & Text          & Book covers, movie covers        & Semi-Automation            & English           \\
ST-VQA \cite{biten2019scene}           & Text          & Multiple sources                 & Human annotation           & English           \\
DocVQA \cite{mathew2021docvqa}          & Text          & Scanned documents, receipts etc. & Human annotation           & English           \\
VisualMRC \cite{tanaka2021visualmrc}       & Text          & Internet sources                 & Human annotation           & English           \\ \hline
\end{tabular}
\end{adjustbox}
\end{table*}

The development of Visual Question Answering (VQA) has been driven by several large-scale English datasets, as summarized in Table \ref{en_dataset}, which also informed the design of ViTextVQA. Early benchmarks such as DAQUAR \cite{malinowski2014multi} initiated the Visual Turing Challenge by evaluating models' multimodal reasoning. VQA v1 \cite{antol2015vqa} introduced open-ended questions and inspired numerous methods (e.g., MCB, MUTAN, NMN), while VQA v2 \cite{goyal2017making} addressed dataset bias by pairing similar images with different answers.

Scene text understanding became a key focus with datasets like TextVQA 
 \cite{singh2019towards}, which introduced 45,336 OCR-based questions and the LoRRA model. OCR-VQA \cite{mishra2019ocr} expanded this to over 1 million QA pairs from 207,572 book covers. ST-VQA \cite{biten2019scene} further improved this area by requiring models to reason over text in natural images.

Beyond natural scenes, DocVQA \cite{mathew2021docvqa} tackled document understanding with 50,000 QA pairs from invoices, forms, and contracts. VisualMRC \cite{tanaka2021visualmrc} advanced this with over 30,000 abstractive QA pairs from 10,000 diverse document images, requiring more complex language generation and document comprehension.

These datasets collectively highlight the evolution of VQA from general image-question tasks to specialized domains involving OCR and document analysis, guiding the creation of ViTextVQA.

\subsubsection{Visual Question Answering Datasets in Vietnamese}
\label{vi_dataset_sec}

To develop ViTextVQA, prior Vietnamese VQA datasets (Table~\ref{vi_dataset}) are extended to better capture Vietnamese linguistic and cultural nuances. The ViVQA dataset \cite{tran2021vivqa} was the first of its kind but lacks natural language fluency and excludes text in images. EVJVQA \cite{nguyen2023vlsp} provides 33,790 multilingual question-answer pairs from Vietnamese images, yet it overlooks scene text. OpenViVQA \cite{nguyen2023openvivqa} introduces open-ended answers with 37,914 pairs but is less effective for classification tasks. ViOCRVQA \cite{pham2025viocrvqa} focuses on OCR-VQA with 120,000 pairs from book covers but lacks image diversity. These limitations underscore the need for a more comprehensive and culturally relevant Vietnamese VQA dataset, which ViTextVQA aims to fulfill by offering several key advantages:
\begin{itemize}
  \item \textbf{Diverse contexts and visual content:} It covers a wide range of real-life domains such as education, healthcare, transportation, culture, and daily life in Vietnam.
  \item \textbf{Integration of scene text and visual cues:} It emphasizes understanding from both textual elements embedded in images and visual scenes, enabling more comprehensive multimodal reasoning.
  \item \textbf{Natural and semantically rich QA pairs:} Questions and answers are curated with fluent Vietnamese language, making them closer to real-world use cases.
  \item \textbf{Cultural and linguistic relevance:} The dataset reflects Vietnamese-specific vocabulary, expressions, and pragmatic usage, facilitating deeper comprehension of local language and context.
\end{itemize}

\begin{table*}[ht]
\centering
\caption{Overview of the former VQA datasets in Vietnamese.}
\label{vi_dataset}
\begin{adjustbox}{width=1\textwidth}
\begin{tabular}{lcccr}
\hline
\textbf{Dataset} & \textbf{Type}    & \textbf{Image Source}        & \textbf{Annotation Method} & \textbf{Language}              \\ \hline
ViVQA \cite{tran2021vivqa}           & Non text         & MS COCO                      & Semi-Automation            & Vietnamese                     \\
EJVVQA \cite{nguyen2023vlsp}          & Non text         & Internet sources             & Human annotation           & English, Japanese, Vietnamese \\
OpenViVQA \cite{nguyen2023openvivqa}       & Text \& Non text & Internet sources             & Human annotation           & Vietnamese                     \\
ViOCRVQA \cite{pham2025viocrvqa}       & Text  & Internet sources - bookcover             & Semi-Automation           & Vietnamese                     \\
ViTextVQA(ours)        & Text             & Internet sources, hand-craft & Human annotation           & Vietnamese     \\          \hline     
\end{tabular}
\end{adjustbox}
\end{table*}

\subsection{Visual Question Answering Methods}
\label{related_methods}
In this study, VQA models are classified into three main groups: CNN-RNN, CNN-LM, and ViT-LM, based on prior research works \cite{wang2016cnn, rafiepour2023ctran, hu2024matryoshka}.

\subsubsection{CNN-RNN Based Methods}
CNN-RNN is one of the earliest VQA approaches, combining CNNs for image feature extraction and RNNs for language understanding. Representative models include VIS+LSTM \cite{strobelt2017lstmvis} using VGGNet \cite{simonyan2014very} and LSTM \cite{hochreiter1997long}, SMem-VQA \cite{xu2016ask} with GoogLeNet \cite{szegedy2015going} and BoW \cite{zhang2010understanding}, and Up-Down \cite{anderson2018bottom} with ResNet \cite{he2016deep} and GRU \cite{chung2014empirical}. These models integrate visual and textual modalities effectively but struggle with long or complex questions. 
Specifically, the Up-Down model introduced a bottom-up attention mechanism, where object-level features from Faster R-CNN are dynamically attended to by a top-down RNN controller, enabling more fine-grained reasoning compared to global CNN features.

\subsubsection{CNN-LM Based Methods}
Replacing RNNs with Transformer-based language models (e.g., BERT \cite{devlin2018bert}), CNN-LM methods enhance contextual understanding. Notable examples include ViLBERT \cite{lu2019vilbert}, VisualBERT \cite{li2019visualbert}, Unicoder-VL \cite{li2020unicoder}, LXMERT \cite{tan2019lxmert}, VL-BERT \cite{su2019vl}, UNITER \cite{chen2020uniter}, and OSCAR \cite{li2020oscar}, all of which use CNNs (e.g., Faster R-CNN \cite{girshick2014rich}) and transformer LMs with different fusion techniques to improve performance. 
Among them, LXMERT adopts a dual-stream architecture, where visual and textual features are first encoded separately and then fused through cross-modal attention layers. UNITER, on the other hand, uses a single-stream transformer that jointly encodes concatenated image-region and text embeddings, achieving stronger alignment across modalities.

\subsubsection{ViT-LM Based Methods}
Transformers have also advanced computer vision with ViT \cite{dosovitskiy2020image}, leading to ViT-LM models that unify visual and textual transformers. Prominent models include BLIP \cite{li2022blip}, BLIP-2 \cite{li2023blip}, GIT \cite{wang2022git}, mPlug \cite{li-etal-2022-mplug}, Flamingo \cite{alayrac2022flamingo}, BeiT-3 \cite{wang2023image}, and PaLI \cite{chen2022pali}. These models achieve state-of-the-art results but require extensive computation and training time. 
For instance, BLIP-2 introduces a query transformer to connect frozen image encoders with large language models, enabling efficient pre-training without fine-tuning the LLM backbone. Flamingo, developed by DeepMind, integrates a visual encoder with a frozen LLM through gated cross-attention layers, allowing the model to handle multi-image and multi-modal in-context learning scenarios.

For further details on resource usage of CNN-LM and ViT-LM models on the ViTextVQA dataset, refer to Section \ref{sec5}.


    

\section{Dataset}\label{sec3}
\subsection{Dataset Creation}
\label{sec3.1}
\begin{figure}[htp]
    \centering
    \includegraphics[width=1\textwidth]{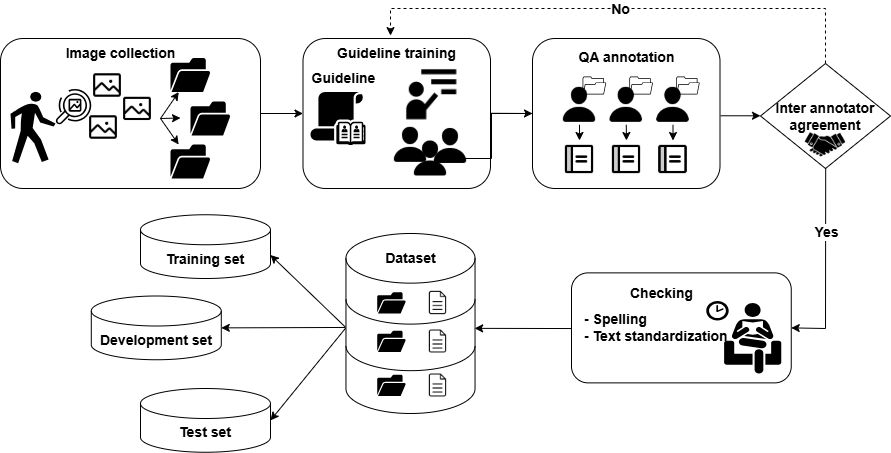}
    \caption{The overview process of creating the ViTextVQA dataset.}
    \label{data_creation}
\end{figure}

The creation process of the ViTextVQA dataset is illustrated in Figure \ref{data_creation}. More than 40,000 images were collected from various open sources such as Google\footnote{\url{https://www.google.com/}}, Facebook\footnote{\url{https://www.facebook.com/}}, Pinterest\footnote{\url{https://www.pinterest.com/}}, and through manual photography. To ensure text relevance, the images were filtered by using SwinTextSpotter \cite{huang2022swintextspotter}, retaining over 22,000 high-quality Vietnamese text-rich images. Subsequently, around 16,000 carefully selected images were manually annotated, resulting in approximately 50,000 question-answer (QA) pairs that form the final ViTextVQA dataset.

\subsubsection{Image Collection}
A keyword list related to major Vietnamese cities (e.g., ``bảng hiệu ở Thành phố Hồ Chí Minh'', ``khu du lịch ở Hội An'') was built to crawl images using Selenium\footnote{\url{https://www.selenium.dev/}} from platforms like Google Images and Pinterest. Manual collection and local photography were also used to enhance diversity. After filtering, the images were organized into folders for annotation.

\subsubsection{Question-Answer Annotation}
Detailed annotation guidelines were defined to ensure that each question is based on visible text in the image, and answers are text spans within the image. Key rules include: no yes/no, location, color, or counting questions; only lowercase answers; and one image having up to 10 QA pairs. Annotators were required to follow consistency rules (e.g., format, spelling) and avoid ambiguous or repetitive content.

\subsubsection{Annotation Process}
\begin{figure}[htp]
\centering
\includegraphics[width=0.75\textwidth]{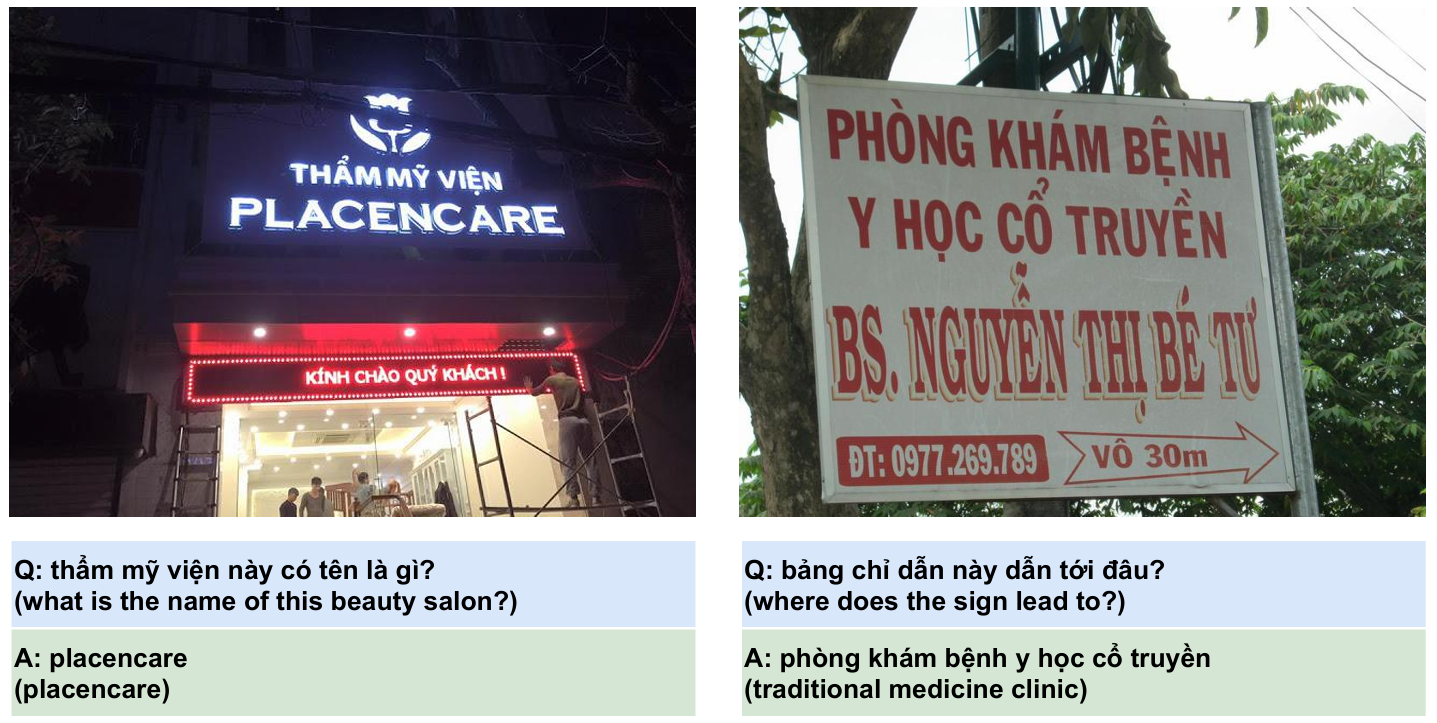}
\caption{Images and QA pairs extracted from the ViTextVQA dataset.}
\label{example_data_2}
\end{figure}
A total of 16 trained undergraduate annotators were recruited. Each annotator created at least 600 QA pairs per subset. A team member supervised every 4 annotators to ensure quality. Annotators were trained on ~100 sample images and reviewed periodically. Quality control included spelling checks, format normalization, and supervisor validation. After six weeks, over 50,000 question-answer pairs were created. The dataset was randomly split into training, development, and test sets using a 7:1:2 ratio. Figure~\ref{example_data_2} illustrates examples of images with their corresponding question-answer pairs from the dataset.

\subsubsection{Inter-Annotator Agreement}
\label{Inter-Annotator Agreement}
To ensure annotation consistency, Inter-Annotator Agreement (IAA) was measured biweekly using an F1-score-based method \cite{rajpurkar-etal-2016-squad}, which is suitable for open-ended QA without fixed label sets. Annotators' answers were compared to reference answers provided by supervisors on 100 samples. The F1-score is calculated based on overlapping words between predictions and ground truth, as detailed in Section \ref{f1}. Annotators were required to achieve an F1 score above 70\%.

\begin{table}[htp]
\centering
\caption{Results of two rounds of Inter-Annotator Agreement measurements.}
\label{tabel1}
\scalebox{1}{
\setlength{\tabcolsep}{10pt}
\begin{tabular}{rcc} 
\toprule
 & \textbf{First Time} &  \textbf{Second Time} \\ 
\midrule
min F1-score (\%) & 73.91 & 83.36 \\
max F1-score (\%) & 87.73 & 87.05 \\
avg F1-score (\%) & 81.83 & 85.34 \\ 
\bottomrule
\end{tabular}
}
\end{table}

After measuring the Inter-Annotator Agreement through the F1-score twice, the results are summarized in Table \ref{tabel1}. In both evaluations, the lowest F1-score values are relatively high (73.91 and 83.36), indicating a considerable level of uniformity between annotators. In addition, the average F1-score increased from the first to the second evaluation, from 81.83 to 85.34, reflecting an improvement in consistency among annotators. These results suggest that the annotated dataset has improved in quality over sessions, and the measurement can be considered reliable for confirming the overall dataset quality.

\subsection{Dataset Analysis}
\label{sec3.2}
\begin{figure}[!ht]
    \centering
    \includegraphics[width=0.95\textwidth]{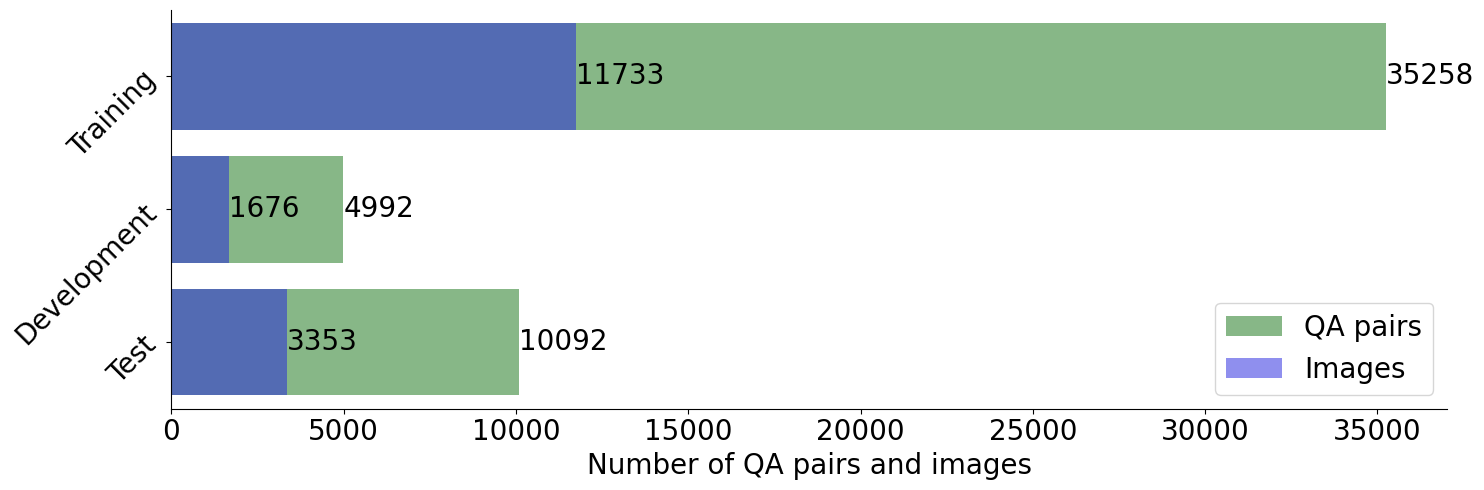}
    \caption{Statistic of images and question-answer pairs.}
    \label{num_ann_img}
\end{figure}

\begin{table}[htp]
\centering
\caption{Comparison between well-known VQA datasets.}
\label{wellnkown-dataset}
\scalebox{0.9}{
\begin{tabular}{lcrrr}
\toprule
\textbf{Dataset} & \textbf{Language} & \textbf{Images} & \textbf{Questions} & \textbf{Answers} \\
\midrule
TextVQA \cite{singh2019towards}      & English     & 28,408  & 45,336  & 453,360 \\
ST-VQA \cite{biten2019scene}         &             & 23,038  & 31,791  & 31,791  \\
DocVQA \cite{mathew2021docvqa}       &             & 12,767  & 50,000  & 50,000  \\
VisualMRC \cite{tanaka2021visualmrc} &             & 10,197  & 30,562  & 30,562  \\ \hline
VAQA \cite{kamel2023vaqa} &         Arabic   & 5,000  & 137,888  & 137,888  \\ \hline
BVQA \cite{10878995} &         Bangla   & 3,545  &  17,725 & 17,725  \\ \hline
FM-IQA \cite{gao2015you} &         Chinese   & 158,392  & 316,193  & 316,193  \\ \hline
ViVQA \cite{tran2021vivqa}           & Vietnamese  & 10,328  & 15,000  & 15,000  \\
OpenViVQA \cite{nguyen2023openvivqa} &             & 11,199  & 21,271  & 21,271  \\
ViCLEVR \cite{tran2023viclevr}       &             & 26,000  & 30,000  & 30,000  \\
\textbf{ViTextVQA (Ours)}            &             & \textbf{16,762} & \textbf{50,342} & \textbf{50,342} \\
\bottomrule
\end{tabular}
}
\end{table}

The ViTextVQA dataset includes 16,762 images accompanied by 50,342 QA pairs. Figure \ref{num_ann_img} depicts the size of training, development and test sets in the dataset. This dataset offers a diverse range of visual scenes as well as scene texts and corresponding questions and answers. Each image is annotated with some question-answer pairs, providing valuable insights into scene text and object required for accurate model predictions where the answer cannot be outside OCR text in the image and this makes the dataset more special in the VQA task. Moreover, statistical comparisons with other widely used datasets in this field are provided in Table \ref{wellnkown-dataset}.

\subsubsection{Question Length}
\begin{figure}[!ht]
    \centering
    \includegraphics[width=0.9\textwidth]{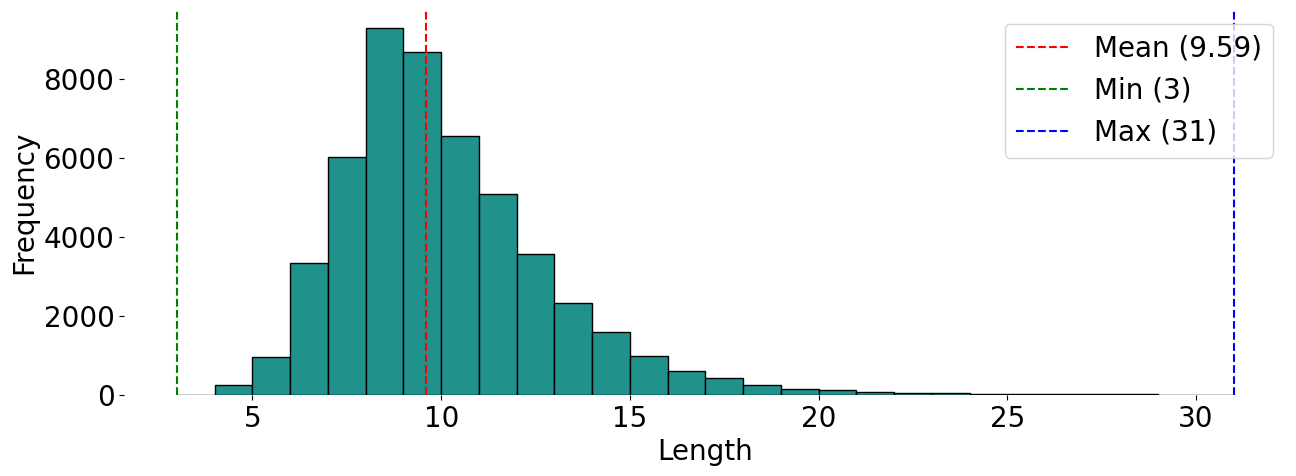}
    \caption{Distribution of question length.}
    \label{ques_len}
\end{figure}

The question length statistics are shown in Figure \ref{ques_len}. Question lengths in the dataset vary significantly, ranging from very short (3 tokens) to quite long (31 tokens). This diversity reflects a wide spectrum of question complexity. Short questions typically target basic information, while longer ones often involve detailed or multi-faceted content. The average question length is 9.59 tokens, suggesting that most questions are concise yet informative, balancing clarity with sufficient context.

\newpage
\subsubsection{Answer Length}
\label{Answer Analysis}

\begin{figure}[!ht]
    \centering
    \includegraphics[width=1\textwidth]{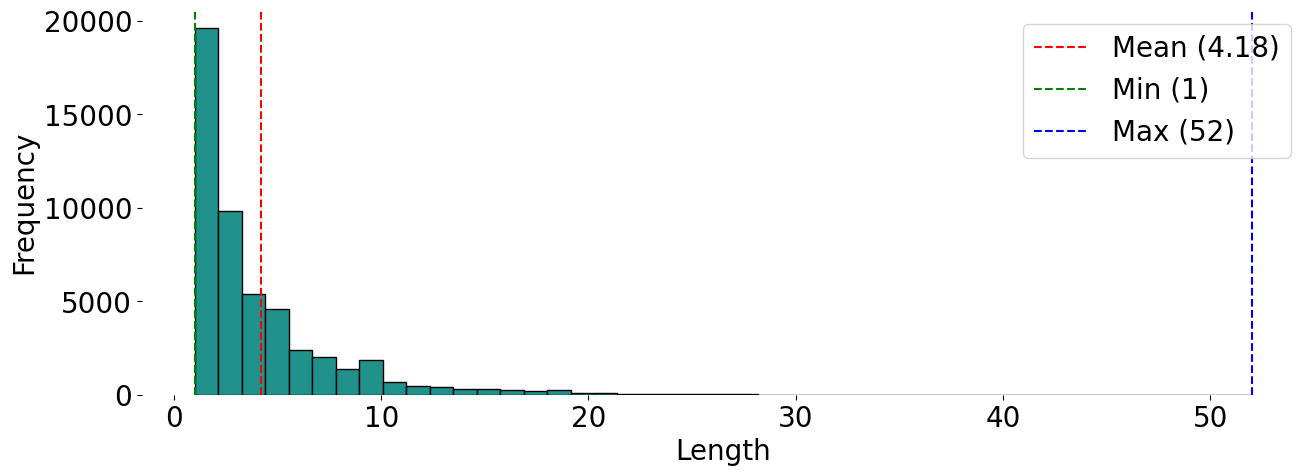}
    \caption{Distribution of answer length.}
    \label{ans_len}
\end{figure}

Figure \ref{ans_len} illustrates the variation in answer lengths, ranging from very short (1 token) to as long as 52 tokens. Despite this wide range, the average answer length is only 4.18 tokens, indicating that most answers are concise. This is primarily due to the annotation constraint that answers must consist solely of text present in the image, resulting in a high frequency of answers from 1 and to tokens.

\subsubsection{Object in Image}

Using VinVL \cite{zhang2021vinvl}, detailed object information such as names, locations, and attributes was extracted, enhancing visual understanding in the ViTextVQA dataset.

\begin{figure*}[htp]
    \centering
    \includegraphics[width=1\textwidth]{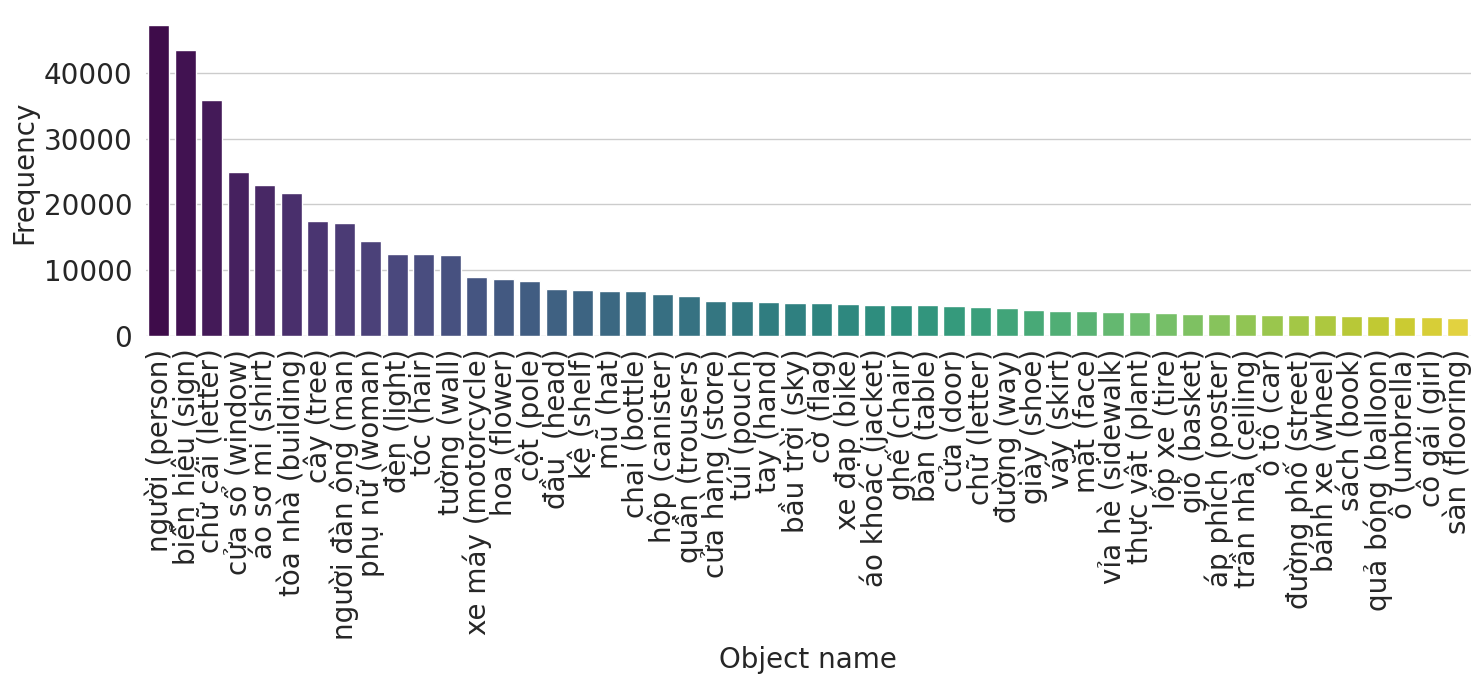}
    \caption{Distribution of the 50 most frequent objects in images.}
    \label{fre_obj}
\end{figure*}

Figure \ref{fre_obj} presents the frequency distribution of the top 50 object categories. ``Person'' is the most frequent object, appearing over 47,000 times, highlighting the central role of humans in Vietnamese daily life scenes. ``Sign'' and ``letter'' follow with 43,000 and 36,000 instances, respectively, reflecting the dataset’s emphasis on scene text and signage-key elements in text-based VQA.

Beyond textual elements, the dataset includes diverse visual contexts, with frequent objects such as ``building'', ``tree'', ``street'', and ``sidewalk''. Clothing-related categories like ``shirt'', ``hat'', and ``jacket'' also appear frequently, indicating relevance to fashion and lifestyle.

Interestingly, ``motorbike'' appears nearly 9,000 times-about three times more than ``car'' (over 3,000 instances), mirroring Vietnam's urban mobility patterns and cultural landscape. This object distribution supports the dataset's contextual richness and cultural alignment.

\subsection{Comparison with Other Visual Question Answering Datasets}

To better understand the characteristics of ViTextVQA, comparisons are made with several popular VQA datasets based on two key textual features: question length and answer length. These factors directly affect model performance. Short questions are often easier to process, while longer ones provide richer context and require deeper reasoning. Likewise, longer answers allow for more informative and expressive responses.

\begin{figure}[!ht]
    \centering
    \includegraphics[width=0.94\textwidth]{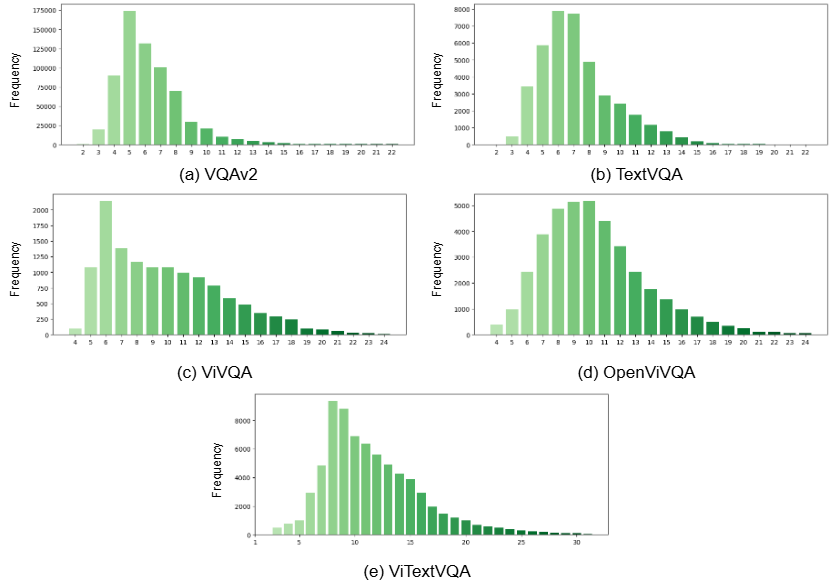}
    \caption{Comparison of question length among VQA datasets.}
    \label{question_length_comparison}
\end{figure}

\begin{figure}[!ht]
    \centering
    \includegraphics[width=0.94\textwidth]{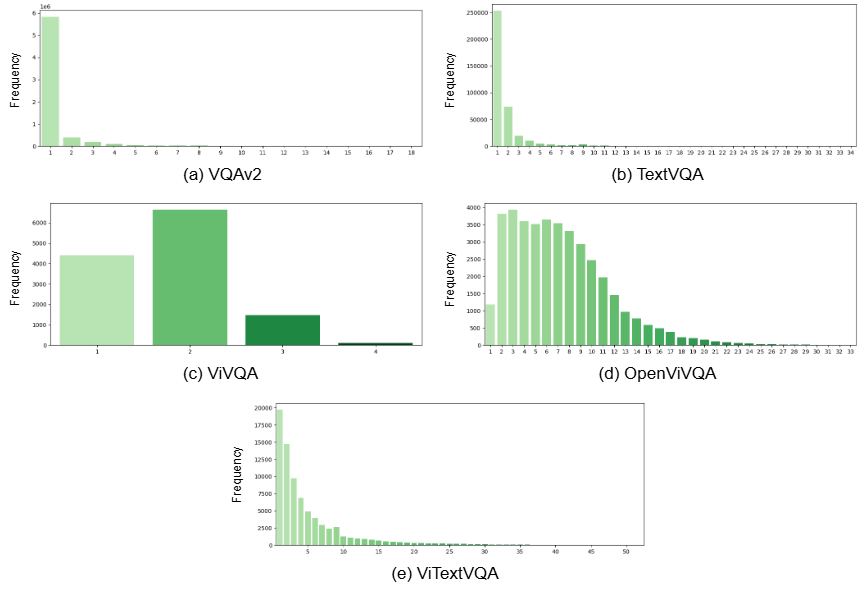}
    \caption{Comparison of answer length among VQA datasets.}
    \label{answer_length_comparison}
\end{figure}

As shown in Figure \ref{question_length_comparison}, ViTextVQA contains a higher proportion of long questions compared to VQAv2, TextVQA, and ViVQA, which mainly include short queries. ViTextVQA consistently features questions longer than 15 tokens, making it suitable for tasks that demand detailed understanding. While OpenViVQA also offers diverse question lengths, ViTextVQA includes more high-length examples, increasing its potential for complex reasoning.

Figure \ref{answer_length_comparison} illustrates that ViTextVQA supports longer answers than VQAv2, ViVQA, and TextVQA. Although OpenViVQA shows a more balanced distribution, ViTextVQA allows responses up to 50 tokens, which enables detailed and comprehensive answers. These properties make ViTextVQA a valuable resource for research in visual question answering that involves both visual understanding and text extraction.

\section{Methodology}\label{sec4}
\begin{figure}[!ht]
    \centering
    \includegraphics[width=1\textwidth]{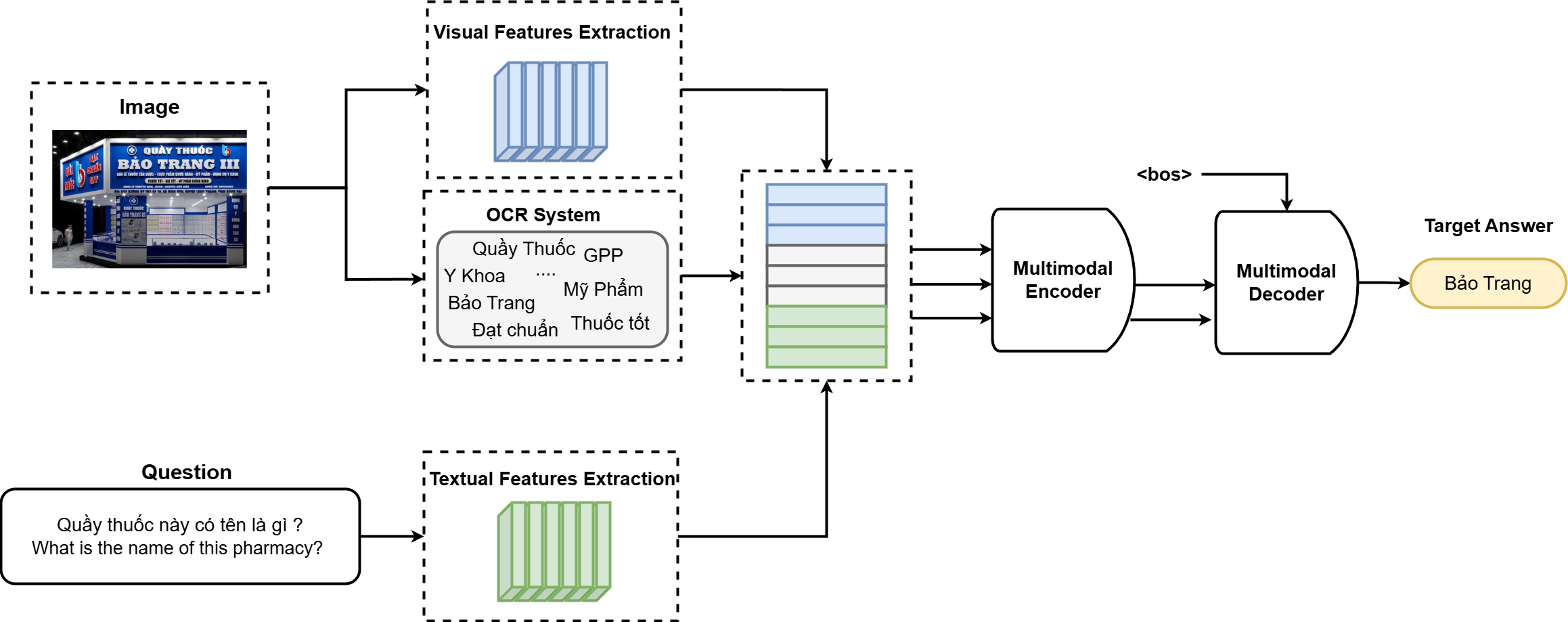}
    \caption{Overview of the structure of text-based VQA models.}
    \label{baseline_models_structure}
\end{figure}

Text-based VQA systems typically consist of three core components: visual feature extraction, textual feature encoding, and an OCR module (Figure \ref{baseline_models_structure}). The visual encoder extracts image features, the textual encoder processes the input question, and the OCR module recognizes embedded scene text. The combination of these components enables the system to reason over both visual and textual modalities for answering questions effectively.

\subsection*{\textbf{ViTextBLIP-2 Architecture}}
\begin{figure}[!ht]
    \centering
    \includegraphics[width=1.0\textwidth]{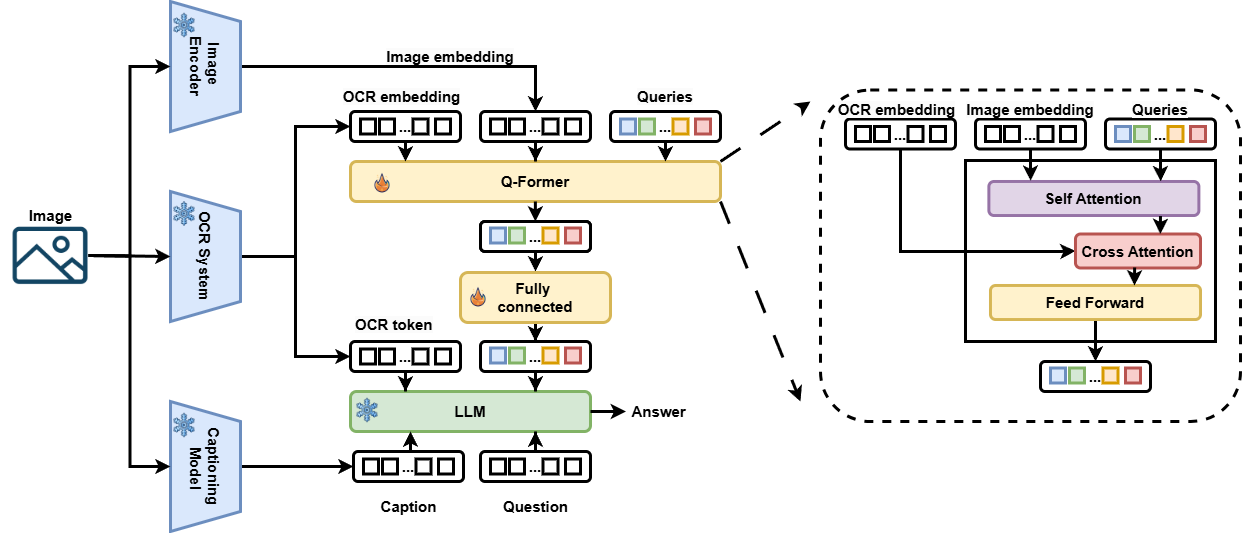}
    \caption{Architecture of ViTextBLIP-2.}
    \label{ViTextBLIP2}
\end{figure}

ViTextBLIP-2 (Vietnamese Text-based Bootstrapped Language-Image Model via Fine-tuning) (Figure \ref{ViTextBLIP2}) is proposed as an adaptation of the BLIP-2 framework, optimized for Vietnamese text-based VQA under limited computational and data constraints. Instead of training the entire model, only the Q-Former is fine-tuned while other modules remain frozen, thereby preserving the pre-trained knowledge of large vision and language models and reducing the risk of overfitting on relatively small Vietnamese datasets. By integrating OCR embeddings and captioning signals, the architecture is particularly effective for text-rich images, enabling robust multimodal reasoning. Concentrating the learning process on the Q-Former further enhances cross-modal alignment between visual, textual, and scene-text features.

The inclusion of OCR and caption information strengthens robustness in real-world scenarios, where text-heavy images are prevalent. Moreover, the modular and lightweight design not only reduces training cost but also facilitates adaptability when combining different image encoders or language models without retraining the entire architecture.

\textbf{Image Encoder:} A pre-trained Vision Transformer (ViT) is adopted to extract rich visual embeddings from the input image:
\begin{equation}
E_{\text{img}} = \text{ViT}(I),
\end{equation}
where \( E_{\text{img}} \) represents image features and \( I \) is the input image.  
Using ViT allows global contextual reasoning across the image and avoids locality limitations of CNNs, which is critical for complex scene understanding in VQA tasks.

\textbf{OCR System:} To handle scene text, SwinTextSpotter \cite{huang2022swintextspotter} is integrated to detect and encode text tokens:
\begin{equation}
\{T_{\text{OCR}}, E_{\text{OCR}}\} = \text{SwinTextSpotter}(I),
\end{equation}
where \( T_{\text{OCR}} \) are recognized tokens and \( E_{\text{OCR}} \) are their embeddings.  
Including OCR embeddings enables the model to answer questions involving text within images, which is common in Vietnamese VQA datasets with signage, menus, or product labels.

\textbf{Captioning Module:} To provide semantic context, Qwen2-VL is employed to generate descriptive captions:
\begin{equation}
C = \text{Qwen2VL}(I),
\end{equation}
where \( C \) is the generated caption.  
Captions help the model by summarizing complex visual scenes into textual form, supporting the LLM in understanding both global context and fine-grained details.

\textbf{Q-Former:} The pre-trained Q-Former, initialized from the original BLIP-2 weights, fuses embeddings from the image encoder, OCR system, and captioning model to support rich contextual reasoning:
\begin{equation}
Q = \text{Q-Former}(E_{\text{img}}, E_{\text{OCR}}, Q_{\text{init}}),
\end{equation}
where \( Q \) is the multimodal representation.  
Q-Former leverages self-attention to integrate information within each modality and cross-attention to connect visual, textual, and OCR features. This design enables rich multimodal reasoning without updating the entire LLM, reducing computational overhead while maintaining performance.

\textbf{Large Language Model:} ViT5 \cite{phan2022vit5}, a Vietnamese pre-trained language model, generates the final answer:
\begin{equation}
A = \text{ViT5}(Q, Q_{\text{ques}}, T_{\text{OCR}}, C),
\end{equation}
where \( Q_{\text{ques}} \) is the question embedding and \( A \) is the predicted answer.  
By providing the LLM with a combination of multimodal embeddings, OCR tokens, and captions, the model can handle complex VQA tasks, including those requiring reasoning over text in images. Fine-tuning only the Q-Former ensures low resource usage while effectively aligning image-text representations with the LLM's knowledge.

\subsection{Feature Extraction}
\label{feature_extraction}

To enhance performance, pre-trained models are utilized for textual, visual, and OCR feature extraction. For text encoding, ViT5 \cite{phan2022vit5}, a Vietnamese encoder-decoder model trained on a large corpus, and mBERT-cased \cite{pires2019multilingual}, a multilingual BERT variant, are adopted. Visual features are extracted using ViT \cite{dosovitskiy2020image}, which divides images into patches for transformer-based processing, and VinVL \cite{zhang2021vinvl}, a Faster R-CNN-based detector trained on diverse objects and attributes. For OCR, SwinTextSpotter \cite{huang2022swintextspotter}, an open-source Vietnamese-capable text spotter using Swin Transformer for joint detection and recognition, is employed to enable accurate scene text extraction tailored to Vietnamese.

\section{Experiment and Results}\label{sec5}
\subsection{Baseline Models}
\label{baseline_models}

To assess the difficulty of the ViTextVQA dataset, several state-of-the-art VQA models were evaluated as baselines, selected based on their historical significance and architectural diversity. These include:

\textbf{M4C} \cite{hu2020iterative}, which introduced a generative VQA framework leveraging BERT-based question embeddings and a Dynamic Pointer Network to integrate OCR and vocabulary tokens;

\textbf{LaTr} \cite{biten2022latr}, an encoder-decoder Transformer based on T5, incorporating layout-aware modules for document understanding, including spatial embeddings and visual feature extraction;

\textbf{PreSTU} \cite{kil2023prestu}, which reorders OCR tokens following their positional layout and applies a T5 architecture pre-trained on scene text image data;

\textbf{BLIP-2} \cite{li2023blip}, which bridges frozen image encoders and large language models using a query transformer for efficient vision-language learning;

\textbf{SaL} \cite{fang2023separate}, which addresses the semantic and spatial structure of OCR text via its Text Semantic Separation and Spatial Circle Position modules.

\begin{table}[htp]
\centering
\caption{Features extraction and OCR system of baseline models.}
\label{baseline_structure}
\scalebox{1}{
\begin{tabular}{lcccc}
\toprule
\textbf{Model} & \textbf{Visual Features} & \textbf{Textual Features} & \textbf{OCR System} & \textbf{Method Type} \\
\midrule
M4C      & VinVL   & mBERT-cased & SwinTextSpotter & CNN-LM \\
BLIP-2   & ViT     & ViT5        & SwinTextSpotter & ViT-LM \\
LaTr     & ViT     & ViT5        & SwinTextSpotter & ViT-LM \\
SaL      & VinVL   & ViT5        & SwinTextSpotter & CNN-LM \\
PreSTU   & ViT     & ViT5        & SwinTextSpotter & ViT-LM \\
\bottomrule
\end{tabular}
}
\end{table}

While these models were originally developed for English-language datasets, they were adapted for Vietnamese while retaining their original architectural designs. Table \ref{baseline_structure} summarizes their key components, including OCR systems, feature extractors, and pre-training strategies, with further implementation details provided in the corresponding sections.

\subsection{Large Language Models (LLMs)}
Recently, large language models (LLMs) such as GPT-4o \cite{achiam2023gpt}, Gemini 1.5 Flash \cite{team2023gemini}, and Qwen2-VL \cite{Qwen-VL} have significantly advanced the VQA task. GPT-4o demonstrates strong performance in object recognition, OCR, and chart interpretation. Gemini 1.5 Flash processes text and images in real time, enabling fast and accurate visual reasoning. Qwen2-VL supports complex multilingual OCR and chart analysis using techniques like Naive Dynamic Resolution.

\subsection{Experimental Configuration}
\label{config}
The baseline models were trained on a local machine equipped with an RTX 4090 GPU (24GB) using the Adam optimizer \cite{kingma2014adam}. Training was conducted over 10 epochs, taking approximately 6-7 hours for base models (7h for ViT-LM, 6h for CNN-LM) and 9-10 hours for large models (10h for ViT-LM, 9h for CNN-LM). Hyperparameters included a learning rate of $3 \times 10^{-5}$, dropout rate of 0.2, batch size of 16, and weight decay of $1 \times 10^{-4}$. All experiments were run locally on Ubuntu 20.04 with Python 3.10, PyTorch 2.0.1, and Transformers 4.34.0.

\subsection{Evaluation Metrics}

\subsubsection{Exact Match (EM)}
The Exact Match metric calculates the percentage of predictions that match the ground truth answers exactly, providing a strict measure of correctness. It is defined as:
\begin{equation}
\text{EM} = \frac{N_{\text{exact}}}{N_{\text{total}}},
\end{equation}
where \(N_{\text{exact}}\) is the number of exact matches and \(N_{\text{total}}\) is the total number of samples.

\subsubsection{F1-score}
\label{f1}
F1-score evaluates the overlap between predicted and reference answers at the token level, balancing both precision and recall:
\begin{equation}
\text{Precision} = \frac{N_{\text{correct}}}{N_{\text{predicted}}}, \quad
\text{Recall} = \frac{N_{\text{correct}}}{N_{\text{gold}}}
\end{equation}
\begin{equation}
\text{F1-score} = \frac{2 \times \text{Precision} \times \text{Recall}}{\text{Precision} + \text{Recall}},
\end{equation}
where \(N_{\text{correct}}\) is the number of overlapping tokens, \(N_{\text{predicted}}\) is the number of tokens in the prediction, and \(N_{\text{gold}}\) is the number of tokens in the reference answer.

\subsection{Main Results}
Table~\ref{Table:main_results} presents the performance of various models evaluated on the ViTextVQA test set using F1-score and EM metrics. Among the baseline methods, the large version of PreSTU (ViT-LM) achieves the best results with an F1-score of 44.93\% and an EM of 22.64\%. Other competitive models include SaL and LaTr in both base and large configurations, showing that combining visual and textual features with spatial awareness can enhance performance. Overall, the ViT-LM-based approaches generally outperform CNN-LM-based ones, especially in larger versions with higher model capacities (up to $\approx$900M parameters) and full parameter tuning.

Remarkably, the proposed ViTextBLIP-2 model outperforms all baselines by a significant margin, reaching an F1-score of 53.95\% and an EM of 25.48\%. Despite having a larger overall size ($\approx$2.5B parameters), ViTextBLIP-2 only fine-tunes a small portion of its parameters ($\approx$5\%), specifically the Q-Former module. This lightweight tuning strategy, combined with its effective multimodal design, enables the model to capture both visual semantics and textual information in Vietnamese more efficiently. The results highlight the potential of ViTextBLIP-2 as a strong foundation for advancing VQA research in low-resource language settings.

\begin{table}[ht]
\centering
\caption{Results of models on the ViTextVQA test set.}
\label{Table:main_results}
\scalebox{0.75}{
\begin{tabular}{l c c c c c c}
\toprule
\textbf{Model} & \textbf{Version} & \textbf{Method Type} & \textbf{Model Size} & \textbf{\% Param Tuning} & \textbf{F1-score (\%)} & \textbf{EM (\%)} \\ 
\midrule
M4C & base & CNN-LM & $\approx$140M & $\approx$100\% & 25.46 & 10.26 \\ 
\midrule
BLIP-2 & base & ViT-LM & $\approx$450M & $\approx$25\% & 36.87 & 13.59 \\ 
BLIP-2 & large & ViT-LM & $\approx$1.4B & $\approx$8\% & 38.47 & 15.58 \\ 
\midrule
LaTr & base & ViT-LM & $\approx$330M & $\approx$100\% & 40.41 & 17.85 \\ 
LaTr & large & ViT-LM & $\approx$900M & $\approx$100\% & 42.34 & 20.94 \\ 
\midrule
SaL & base & CNN-LM & $\approx$270M & $\approx$100\% & 43.49 & 20.11 \\ 
SaL & large & CNN-LM & $\approx$780M & $\approx$100\% & 43.39 & 21.48 \\ 
\midrule
PreSTU & base & ViT-LM & $\approx$315M & $\approx$100\% & 43.81 & 20.85 \\ 
PreSTU & large & ViT-LM & $\approx$860M & $\approx$100\% & 44.93 & 22.64 \\ 
\midrule
\textbf{ViTextBLIP-2 (Ours)} & \textbf{base} & \textbf{ViT-LM} & $\approx$\textbf{2.5B} & $\approx$\textbf{5\%} & \textbf{53.95} & \textbf{25.48} \\ 
\bottomrule
\end{tabular}
}
\end{table}

\subsection{Performance of Large Language Models Compared to Human}

In this experiment, LLMs were evaluated on a random subset of 100 samples using 0-shot and few-shot (1, 3, and 5-shot) settings, and the results were compared with human performance.

\begin{table}[ht]
\centering
\caption{Performance of LLMs and humans on 100 randomly sampled ViTextVQA test samples.}
\label{LLM_results}
\scalebox{0.9}{
\begin{tabular}{lcccccccc}
\toprule
\multirow{2}{*}{\textbf{Model}} & \multicolumn{4}{c}{\textbf{F1-score (\%)}} & \multicolumn{4}{c}{\textbf{EM (\%)}} \\ 
\cmidrule(lr){2-5} \cmidrule(lr){6-9} 
& 0-shot & 1-shot & 3-shot & 5-shot & 0-shot & 1-shot & 3-shot & 5-shot \\ 
\midrule
GPT-4o               & 55.51 & \textbf{69.55} & \textbf{76.67} & \textbf{75.62} & 25.00 & \textbf{45.00} & \textbf{56.00} & \textbf{55.00} \\ 
Gemini-1.5-flash     & \textbf{57.75} & 59.09 & 62.37 & 64.40 & 26.00 & 29.00 & 35.00 & 39.00 \\ 
QwenVL-7b            & 53.80 & 54.10 & 54.85 & 54.74 & \textbf{28.00} & 36.00 & 36.00 & 35.00 \\ 
\midrule
ViTextBLIP-2         & \multicolumn{4}{c}{56.43} & \multicolumn{4}{c}{26.00} \\ 
\midrule
Human                & \multicolumn{4}{c}{\underline{96.42}} & \multicolumn{4}{c}{\underline{88.00}} \\ 
\bottomrule
\end{tabular}
}
\end{table}

As shown in Table \ref{LLM_results}, while GPT-4o and Gemini show performance gains with more shots, all models fall short of human-level accuracy. Humans achieved 96.42\% F1 and 88.00\% EM, whereas LLMs peaked at 76.67\% F1 and 56.00\% EM (GPT-4o, 3-shot). These results highlight the current limitations of LLMs in understanding and reasoning over Vietnamese text in complex visual contexts.

\subsection{Compare with Other Datasets}

To assess generalizability beyond ViTextVQA, ViTextBLIP-2 was evaluated on the OpenViVQA dataset \cite{nguyen2023openvivqa}, which features open-ended visual questions. As shown in Table \ref{openvivqa_results}, ViTextBLIP-2 achieves a CIDEr score of 3.2129 and an average BLEU score of 0.4717, significantly outperforming both traditional models (e.g., M4C, MCAN, LoRRA) and recently proposed methods (e.g., QuMLPAG, MLPAG). These results demonstrate the effectiveness and robustness of the model across diverse VQA benchmarks.

\begin{table}[!ht]
\centering
\caption{Experimental results on OpenViVQA test set, inherited from the study of \mbox{\cite{nguyen2023openvivqa}}.}
\label{openvivqa_results}
\scalebox{0.9}{
\begin{tabular}{l c c c c}
\toprule
\textbf{Model} & \textbf{Version} & \textbf{Method Type} & \textbf{CIDEr} & \textbf{Avg BLEU} \\ 
\midrule
MCAN       & base & ViT-LM     & 1.0613 & 0.1699 \\ 
LoRRA      & base & CNN-RNN    & 0.8005 & 0.1349 \\ 
M4C        & base & CNN-LM     & 1.5073 & 0.2941 \\ 
\midrule
FST        & base & CNN-LM     & 0.6141 & 0.1050 \\ 
QuMLPAG    & base & CNN-LM     & 1.7082 & 0.2651 \\ 
MLPAG      & base & CNN-LM     & 1.6104 & 0.2739 \\ 
\midrule
\textbf{ViTextBLIP-2 (Ours)} & \textbf{base} & \textbf{ViT-LM} & \textbf{3.2129} & \textbf{0.4717} \\ 
\bottomrule
\end{tabular}
}
\end{table}

\section{Results Analysis}\label{sec6}
\subsection{Effect of OCR System Performance and Challenges}
\label{effect_performance}

\begin{figure}[!ht]
    \centering
    \includegraphics[height=6cm]{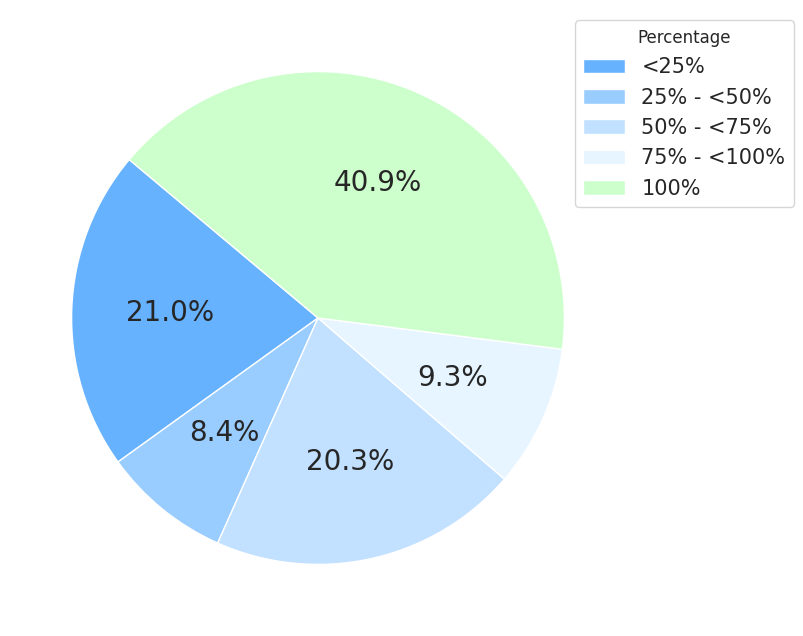}
    \caption{Proportion of answer tokens found in OCR text on the test set.}
    \label{ocr_type_percent}
\end{figure}

The performance of VQA models exhibits a strong correlation with OCR quality. Using SwinTextSpotter, test samples were categorized based on the percentage of answer tokens detected in the OCR output. The categories include: $<$25\%, 25--50\%, 50--75\%, 75--100\%, and 100\% (see Figure \ref{ocr_type_percent}).

\begin{figure}[htp]
  \centering
  \includegraphics[width=0.49\textwidth]{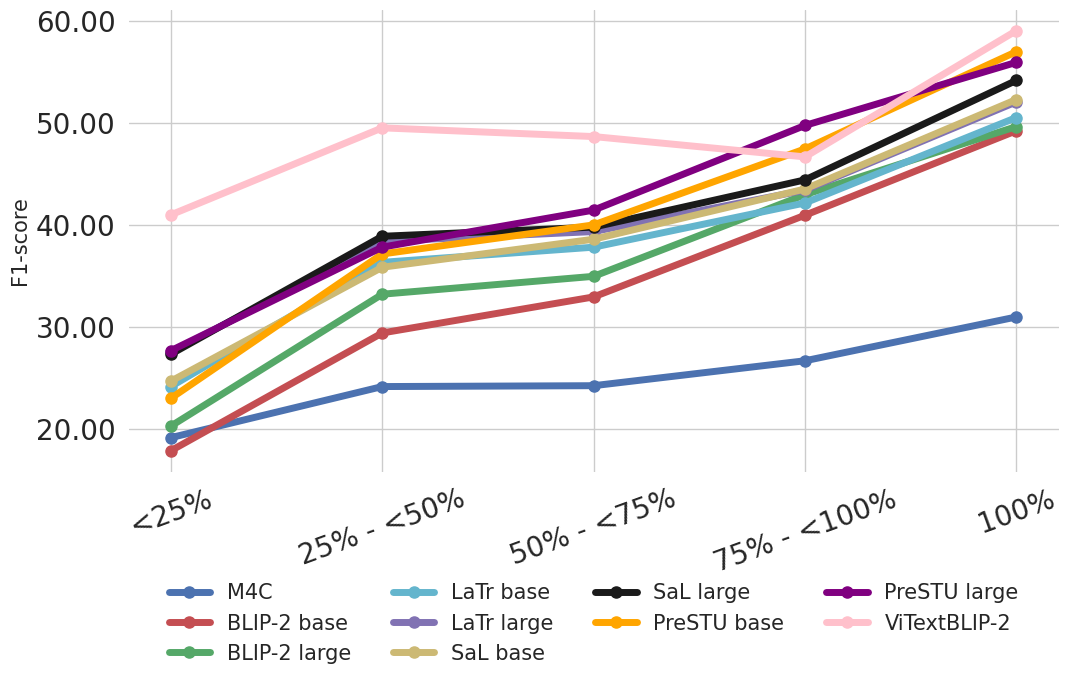}
  \includegraphics[width=0.49\textwidth]{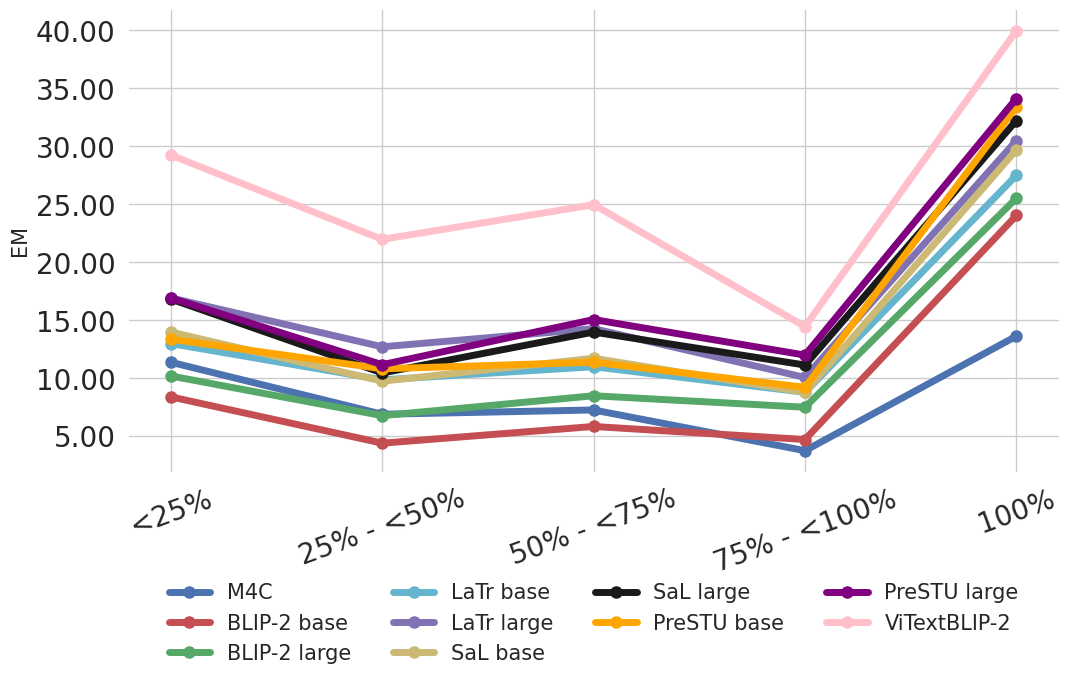}
  \caption{Model performance (F1 and EM) with respect to OCR token coverage.}
  \label{combined_figure_not_sort}
\end{figure}

As shown in Figure \ref{combined_figure_not_sort}, F1-score generally increases with higher OCR token coverage, confirming that OCR-extracted text plays a vital role in VQA prediction. However, EM results fluctuate, with drops at mid-ranges (25-50\% and 75-100\%), then rising sharply at full coverage (100\%).

Despite this, even with full token coverage, the best F1 and EM remain below 60 and 35, respectively. This reflects the challenge of ViTextVQA: models must not only locate correct text spans but also filter noise and integrate visual-textual context effectively.

\subsection{Effect of Answer and Question Length}
\begin{table}[!ht]
  \centering
  \caption{Distribution of answer and question lengths in the test set.}
  \begin{tabular}{lrrr}
    \hline
    \textbf{Group} & \textbf{Length (n)} & \textbf{Answer Samples} &\textbf{Question Samples} \\
    \hline
    Short        & $n \leq 5$        & 7{,}853 & 225 \\
    Medium       & $5 < n \leq 10$   & 1{,}541 & 6{,}834 \\
    Long         & $10 < n \leq 15$  & 421     & 2{,}617 \\
    Very long    & $n > 15$          & 213     & 352 \\
    \hline
  \end{tabular}
  \label{ans_ques_len}
\end{table}

Samples in the test set were grouped by token length (Table \ref{ans_ques_len}) to examine how question and answer length affect model performance. The results reveal clear patterns.

\begin{figure}[!ht]
  \centering
  \includegraphics[width=0.49\textwidth]{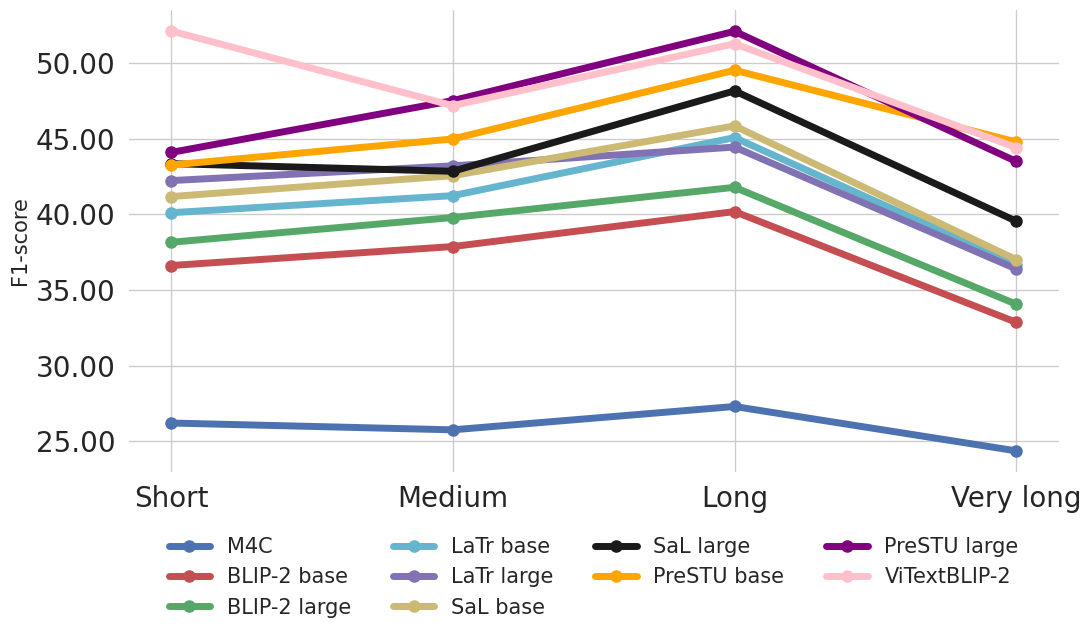}
  \includegraphics[width=0.49\textwidth]{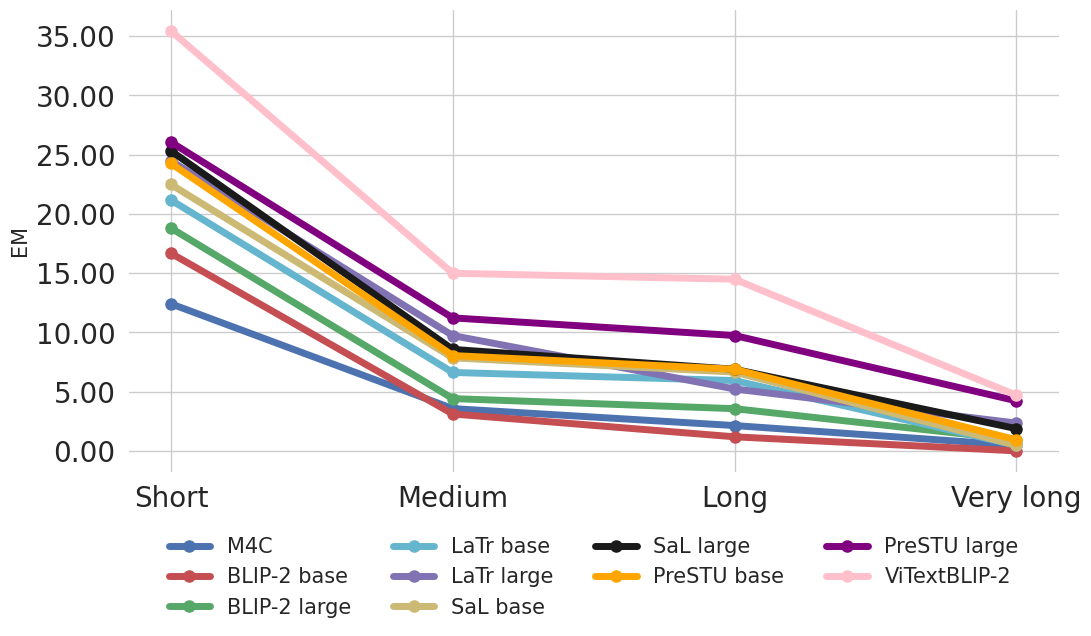}
  \caption{Model performance by answer length.}
  \label{answer length}
\end{figure}

As shown in Figure \ref{answer length}, model performance tends to decline as answer length increases. EM scores drop steadily with longer answers, reflecting reduced exact match capability. For F1-score, medium and long answers yield better performance than short ones, but this trend reverses with very long answers, likely due to accumulated OCR noise or increased linguistic complexity.

\begin{figure}[!ht]
  \centering
  \includegraphics[width=0.49\textwidth]{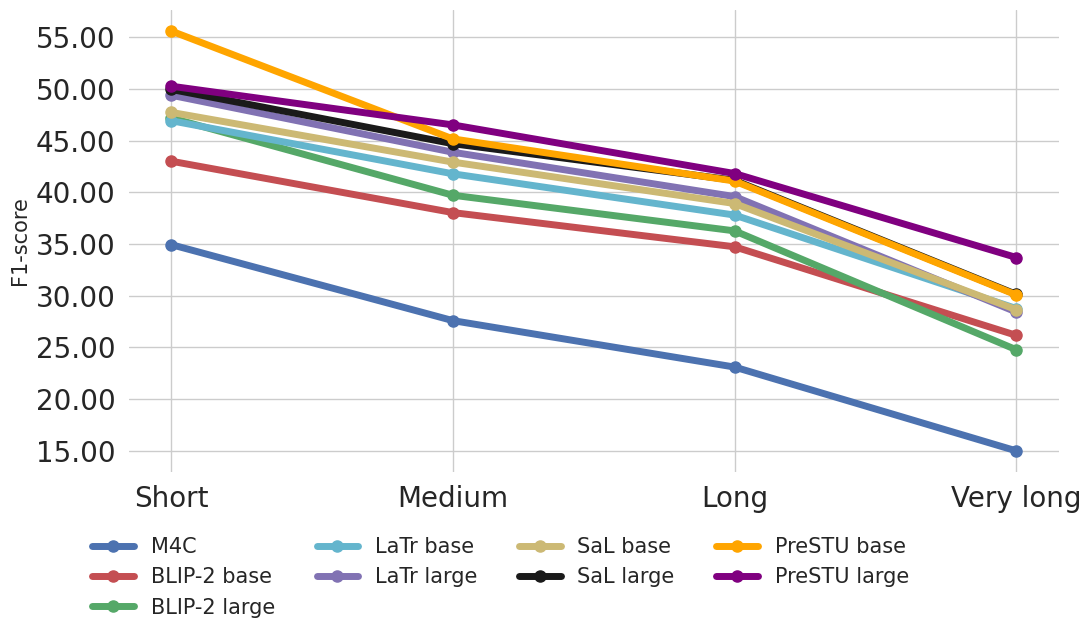}
  \includegraphics[width=0.49\textwidth]{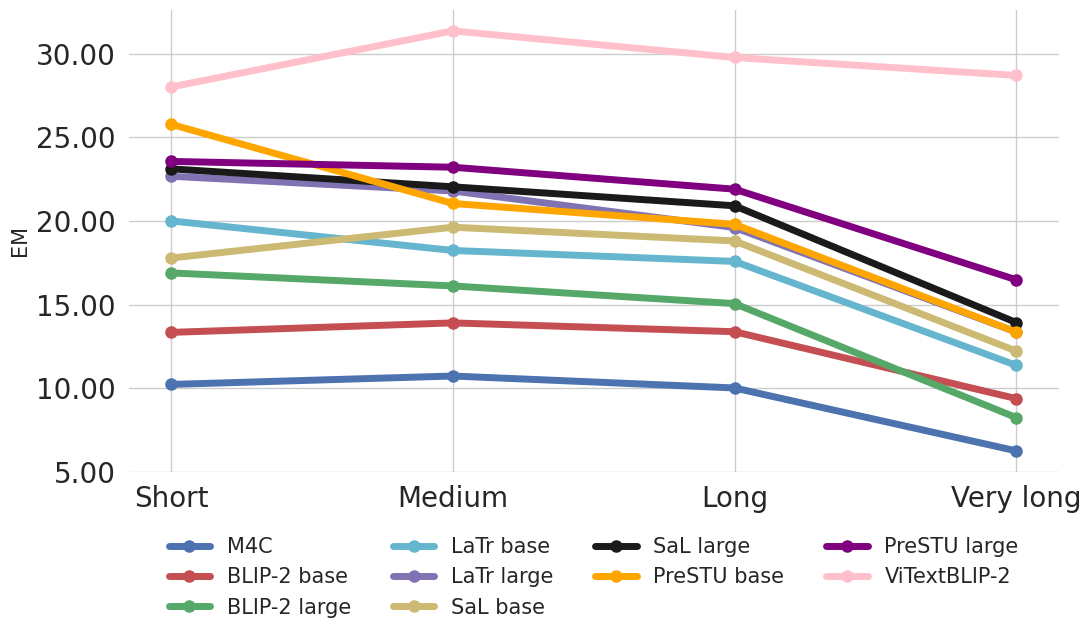}
  \caption{Model performance by question length.}
  \label{question length}
\end{figure}

Regarding question length (Figure \ref{question length}), shorter questions achieve the highest F1 scores. Performance gradually decreases as length increases, with a sharp drop for very long questions. For EM, the performance remains relatively stable across short to long ranges, though individual model behaviors vary. For instance, SaL base improves with question length up to 15 tokens, while PreSTU base declines. Very long questions pose notable challenges for all models, reducing both EM and F1 scores.

\subsection{Effect of OCR Text Size}
\begin{figure}[!ht]
    \centering
    \includegraphics[width=1\textwidth]{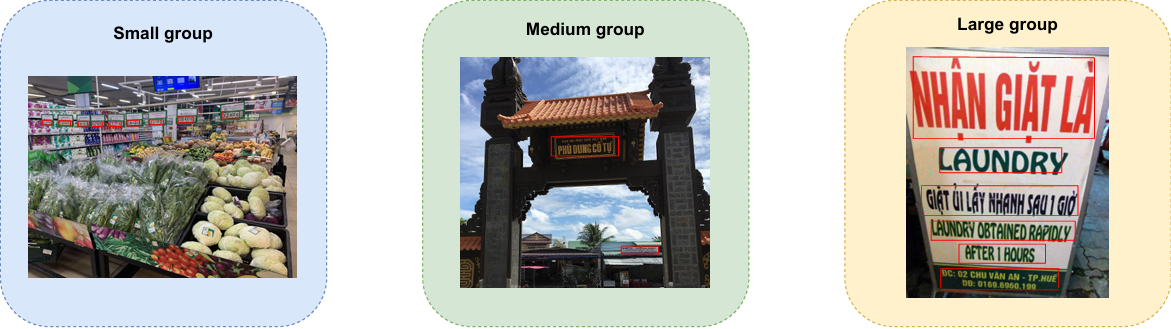}
    \caption{Illustration of OCR text size groups.}
    \label{ocr_size}
\end{figure}

To assess the impact of OCR text size on model performance, only samples in which all answer tokens appeared in the OCR output (as in Section \ref{effect_performance}) were analyzed to ensure reliable measurement. Using VinVL \cite{zhang2021vinvl}, the total bounding-box area of answer tokens was computed as a percentage of the image area. Based on this ratio, samples were grouped as: \textit{small} (<1\%), \textit{medium} (1-5\%), and \textit{large} (>5\%) (Figure \ref{ocr_size}).

\begin{figure}[!ht]
  \centering
  \includegraphics[width=0.49\textwidth]{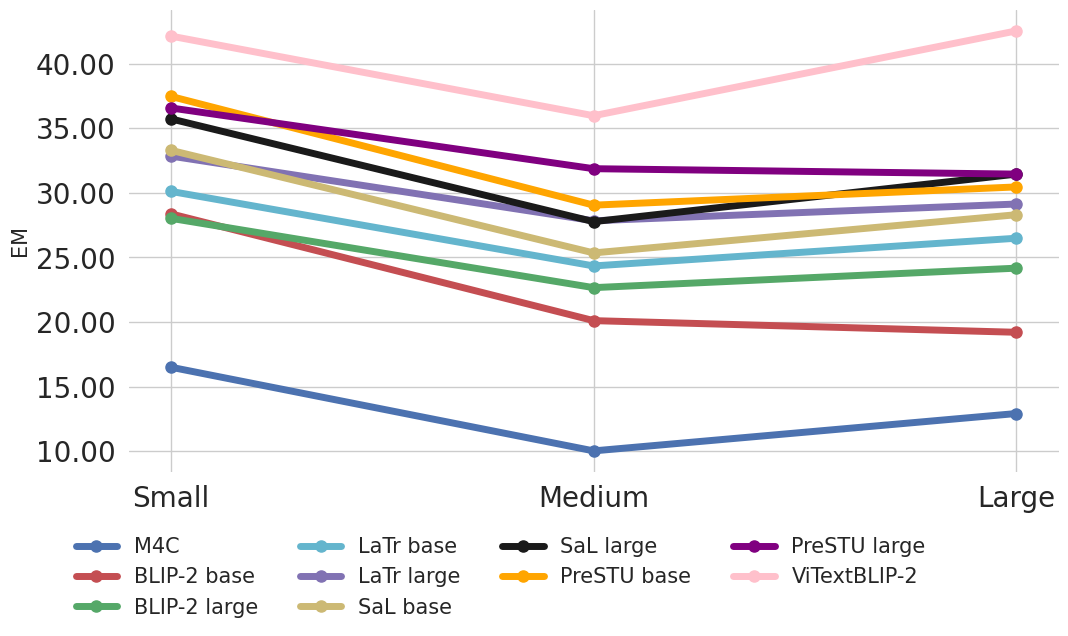}
  \includegraphics[width=0.49\textwidth]{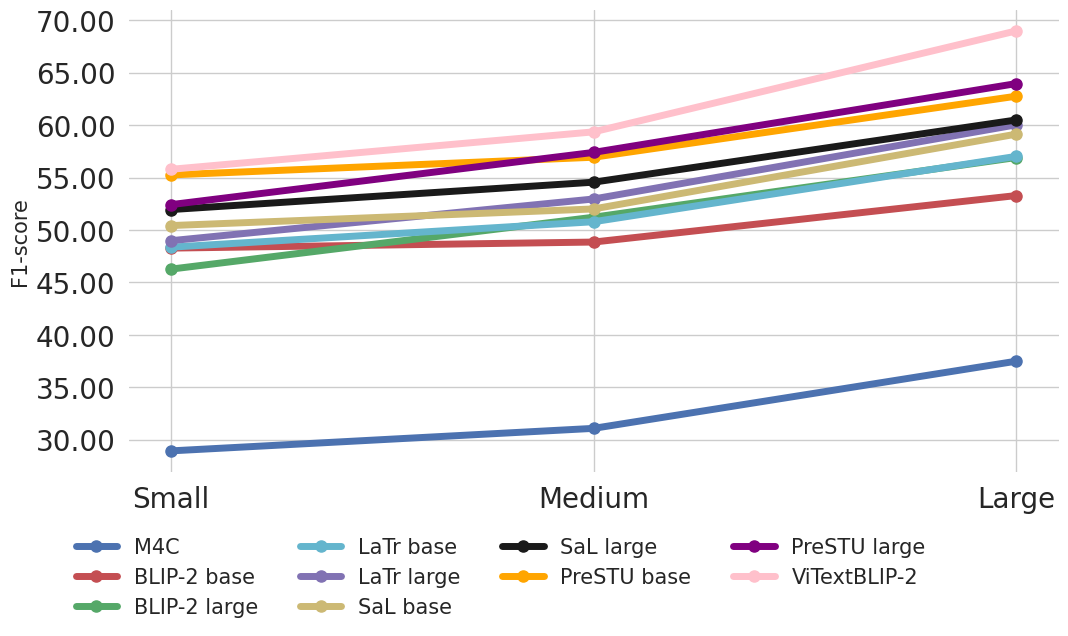}
  \caption{Model performance by OCR text size.}
  \label{combined_figure_origin_size}
\end{figure}

As shown in Figure \ref{combined_figure_origin_size}, F1-score consistently improves with larger OCR text regions across all models, indicating better recognition and contextual grounding. For EM, results are less consistent: performance drops slightly from small to medium sizes but improves again in the large group. This suggests that while larger text regions generally enhance comprehension, exact matching remains sensitive to OCR segmentation quality.

\subsection{Effect of OCR Arrangement}
\label{sec6.5}
In this experiment, the impact of different OCR text arrangements on VQA model performance is investigated. While scene text typically follows a top-left to bottom-right (TL-BR) reading order, OCR outputs may not preserve this structure. Three strategies are therefore evaluated: (1) original order, (2) sorting by OCR confidence (descending), and (3) TL-BR sorting \cite{kil2023prestu}.

\begin{table*}[!ht]
\centering
\caption{Results of models with different OCR arrangements on ViTextVQA dataset. $\triangle$ indicates the increase (↑) or decrease (↓) compared with the original (*).}
\label{ocr_arrangement}
\resizebox{\textwidth}{!}{%
\begin{tabular}{l c c | c c | c c}
\toprule
\multirow{2}{*}{\textbf{Model}} & \multicolumn{2}{c|}{\textbf{ Original Order}} & \multicolumn{2}{c|}{\textbf{OCR
Confidence}} & \multicolumn{2}{c}{\textbf{TL-BR}} \\
\cmidrule(lr){2-7} 
 & \textbf{F1-score (\%)} & \textbf{EM (\%)} & \textbf{F1-score (\%)} & \textbf{EM (\%)} & \textbf{F1-score (\%)} & \textbf{EM (\%)} \\ 
\midrule
M4C & 25.46* & 10.26* & 18.41 {\color{red} (↓7.05)} & 8.29 {\color{red} (↓1.97)} & 30.04 {\color{blue} (↑4.58)} & 11.60 {\color{blue} (↑1.34)} \\ 
BLIP-2 base & 36.87* & 13.59* & 35.21 {\color{red} (↓1.66)} & 12.97 {\color{red} (↓0.62)} & 37.78 {\color{blue} (↑0.91)} & 15.01 {\color{blue} (↑1.42)} \\ 
BLIP-2 large & 38.47* & 15.58* & 36.74 {\color{red} (↓1.73)} & 14.15 {\color{red} (↓1.43)} & 42.16 {\color{blue} (↑3.69)} & 18.56 {\color{blue} (↑2.98)} \\ 
LaTr base & 40.41* & 17.85* & 38.76 {\color{red} (↓1.65)} & 16.32 {\color{red} (↓1.53)} & 43.13 {\color{blue} (↑2.72)} & 20.42 {\color{blue} (↑2.57)} \\ 
LaTr large & 42.34* & 20.94* & 39.72 {\color{red} (↓2.62)} & 18.88 {\color{red} (↓2.06)} & 44.07 {\color{blue} (↑1.73)} & 22.27 {\color{blue} (↑1.33)} \\ 
SaL base & 43.49* & 20.11* & 43.66 {\color{blue} (↑0.17)} & 20.28 {\color{blue} (↑0.17)} & 44.89 {\color{blue} (↑1.40)} & 20.97 {\color{blue} (↑0.86)} \\ 
SaL large & 43.39* & 21.48* & 43.71 {\color{blue} (↑0.32)} & 20.47 {\color{red} (↓1.01)} & 44.74 {\color{blue} (↑1.35)} & 21.18 {\color{red} (↓0.30)} \\ 
PreSTU base & 41.74 {\color{red} (↓2.07)} & 19.15 {\color{red} (↓1.70)} & 40.09 {\color{red} (↓3.72)} & 17.74 {\color{red} (↓3.11)} & 43.81* & 20.85* \\ 
PreSTU large & 42.85 {\color{red} (↓2.08)} & 21.40 {\color{red} (↓1.24)} & 40.13 {\color{red} (↓4.80)} & 18.43 {\color{red} (↓4.21)} & 44.93* & 22.64* \\ 
\midrule
\textbf{ViTextBLIP-2 (Ours)} & 51.16 {\color{red} (↓1.35)} & 24.10 {\color{red} (↓0.73)} & 53.95 {\color{blue} (↑1.44)} & 25.48 {\color{blue} (↑0.65)} & \textbf{52.51*} & \textbf{24.83*} \\ 
\midrule
\textbf{Avg} & 40.62 & 18.45 & 39.04 & 17.30 & 42.81 & 19.83 \\ 
\bottomrule
\end{tabular}%
}
\end{table*}

\begin{table*}[!ht]
\centering
\caption{Results of ViTextBLIP-2 with different OCR arrangements on OpenViVQA dataset. $\triangle$ indicates the increase (↑) or decrease (↓) compared with the original (*).}
\label{ocr_arrangement_openvivqa}
\resizebox{\textwidth}{!}{%
\begin{tabular}{l c c | c c | c c}
\toprule
\multirow{2}{*}{\textbf{Model}} & \multicolumn{2}{c|}{\textbf{ Original Order}} & \multicolumn{2}{c|}{\textbf{OCR
Confidence}} & \multicolumn{2}{c}{\textbf{TL-BR}} \\
\cmidrule(lr){2-7} 
 & \textbf{CIDEr} & \textbf{Avg BLEU} & \textbf{CIDEr} & \textbf{Avg BLEU} & \textbf{CIDEr} & \textbf{Avg BLEU} \\ 
\midrule
\textbf{ViTextBLIP-2 (Ours)} & 2.6329 {\color{red} (↓0.58)} & 0.4125 {\color{red} (↓0.06)} & 3.0024 {\color{red} (↓0.21)} & 0.4502 {\color{red} (↓0.02)} & \textbf{3.2129*} & \textbf{0.4717*} \\ 
\bottomrule
\end{tabular}%
}
\end{table*}

Experiments on both ViTextVQA and OpenViVQA datasets (Tables \ref{ocr_arrangement} and \ref{ocr_arrangement_openvivqa}) reveal consistent trends. Most models benefit from TL-BR sorting, achieving higher F1/EM or CIDEr/BLEU scores. This aligns with Vietnamese reading patterns, enabling models to better preserve textual semantics and sequence.

Conversely, confidence-based sorting generally degrades performance. Rearranging tokens by confidence disrupts the spatial and linguistic coherence of the text, making interpretation more difficult.

Overall, OCR arrangement plays a significant role in model accuracy. Maintaining natural reading order-specifically TL-BR, enhances comprehension, while artificial reordering (e.g., confidence-based) may hinder it.

\newpage
\subsection{Ablation Study}
\begin{table}[!ht]
\centering
\caption{Ablation study on different model components.}
\label{Table:ablation_results}
\scalebox{0.9}{
\begin{tabular}{c c c l l}
\toprule
\textbf{Image Encoder} & \textbf{OCR System} & \textbf{Captioning Model} & \textbf{F1-score (\%)} & \textbf{EM (\%)} \\ 
\midrule
✓ & ✓ & ✓ & 0.5251* & 0.2483* \\ 
✓ & ✓ & ✗ & 0.3914 {\color{red} (↓0.1337)} & 0.1694 {\color{red} (↓0.0789)} \\ 
✓ & ✗ & ✓ & 0.3953 {\color{red} (↓0.1298)} & 0.1794 {\color{red} (↓0.0689)} \\ 
✓ & ✗ & ✗ & 0.1417 {\color{red} (↓0.3834)} & 0.0469 {\color{red} (↓0.2014)} \\ 
✗ & ✓ & ✓ & 0.4167 {\color{red} (↓0.1084)} & 0.1882 {\color{red} (↓0.0601)} \\ 
✗ & ✓ & ✗ & 0.3668 {\color{red} (↓0.1583)} & 0.1332 {\color{red} (↓0.1511)} \\ 
✗ & ✗ & ✓ & 0.3512 {\color{red} (↓0.1739)} & 0.1267 {\color{red} (↓0.1216)} \\ 
✗ & ✗ & ✗ & 0.0016 {\color{red} (↓0.5235)} & 0.0000 {\color{red} (↓0.2483)} \\ 
\bottomrule
\end{tabular}
}
\end{table}

In this section, an ablation study is presented to evaluate the contribution of each component in ViTextBLIP-2: Image Encoder, OCR System, and Captioning Model. The results are summarized in Table \ref{Table:ablation_results}.

\textbf{Full model (✓✓✓):}  
The complete model achieves the best performance (F1: 52.51\%, EM: 24.83\%), confirming that the combination of visual, textual, and semantic features is essential for robust performance.

\textbf{Removing Captioning (✓✓✗):}  
Excluding the captioning module leads to a notable drop (F1: ↓13.37\%, EM: ↓7.89\%), indicating its role in providing complementary semantic cues.

\textbf{Removing OCR (✓✗✓):}  
Without OCR, performance degrades similarly (F1: ↓12.98\%, EM: ↓6.89\%), emphasizing the importance of scene text for text-based VQA.

\textbf{Image Encoder only (✓✗✗):}  
Using only the Image Encoder results in poor performance (F1: 14.17\%, EM: 4.69\%), showing it alone cannot support complex text-based reasoning.

\textbf{No components (✗✗✗):}  
Removing all modules renders the model ineffective (F1: 0.16\%, EM: 0.00\%), highlighting the necessity of multimodal integration.

Overall, the study confirms that each module contributes significantly to ViTextBLIP-2, with OCR and captioning providing critical text and context to support accurate predictions.

\section{Discussion on Characteristics of Language for Vietnamese Text-based Visual Question Answering}
\label{sec7}
For a language such as Vietnamese, unique characteristics must be considered to achieve high performance in the VQA task. This section analyzes and discusses two main aspects of the Vietnamese language: diacritics and word segmentation, as well as their relationship to this task.

\subsection{Impact of Diacritics on Vietnamese Text Understanding}

Vietnamese is a tonal language with a rich diacritic system comprising seven letters (ă, â, ê, ô, ơ, ư, đ) and five tonal marks (à, á, ả, ã, ạ) \cite{Nguyen_Nguyen_Ngo_Vu_Tran_Ngo_Le_2019}. These diacritics are essential for distinguishing phonemes and lexical meanings, e.g.,  ``má '' (mother) vs.  ``mà '' (but). Historically, their development was shaped by Chinese and Nom influences, and later standardized during the colonial era.

To evaluate the role of diacritics, experiments were conducted on the ViTextVQA dataset by progressively removing diacritics at rates of 0\%, 25\%, 50\%, 75\%, and 100\%.

\begin{figure}[htp]
  \centering
\includegraphics[width=0.49\textwidth]{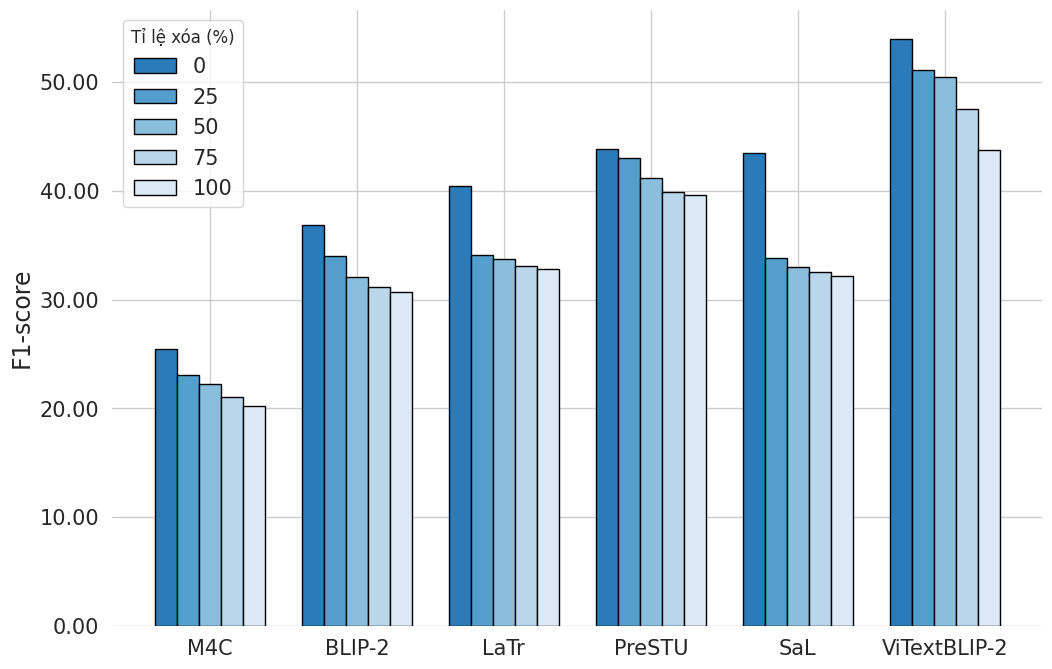}
\includegraphics[width=0.49\textwidth]{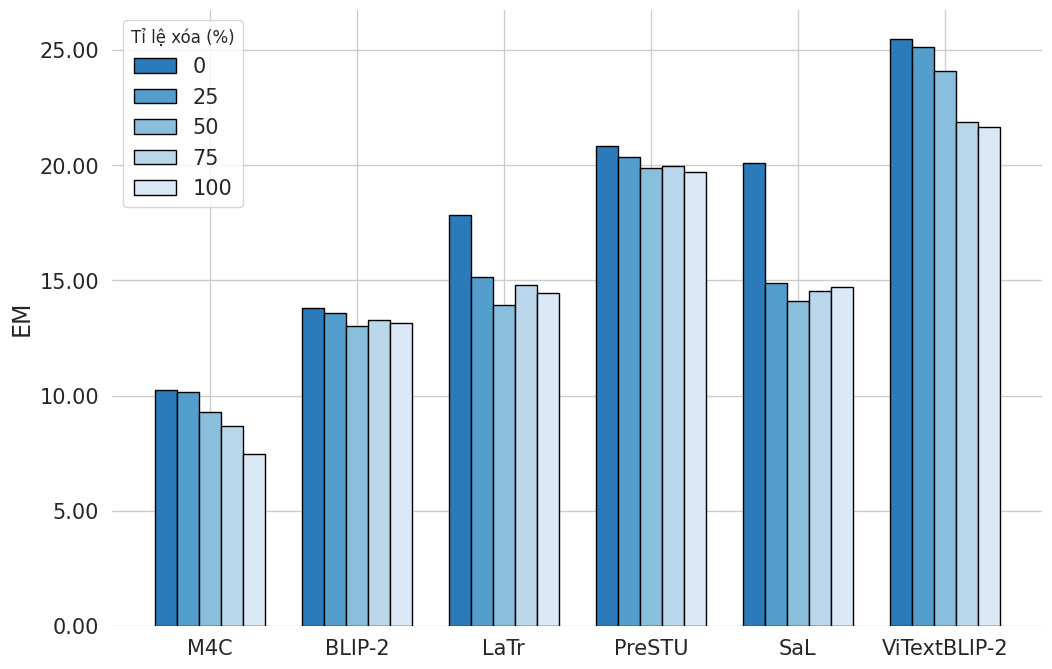}
  \caption{Performance of base models under different diacritic removal rates.}
  \label{diacritics_fig}
\end{figure}

As shown in Figure \ref{diacritics_fig}, removing diacritics leads to a significant performance drop across all models. F1-score decreases consistently with higher removal rates, confirming that diacritics are critical for semantic understanding. EM scores fluctuate more but generally decline, with the sharpest drop at 50\% removal for most models.

Notably, LaTr and SaL are more sensitive to diacritic loss, especially from 0\% to 25\%, suggesting a strong reliance on diacritic information. These findings highlight the importance of diacritics in preserving meaning and improving model comprehension in Vietnamese VQA tasks.

\subsection{Impact of Vietnamese Word Segmentation}

Vietnamese presents unique challenges in word segmentation due to the lack of explicit word boundaries. Unlike English, white spaces in Vietnamese separate syllables, not necessarily words \cite{nguyen-etal-2018-fast}. For example,  ``thời gian '' (time) consists of two syllables forming a single word  ``thời\_gian ''. Accurate segmentation is therefore essential for correct semantic interpretation.

We hypothesize that OCR text arrangements which better preserve the natural reading order of Vietnamese can lead to improved word segmentation and thus better performance. Using VnCoreNLP \cite{vu2018vncorenlp}, segmented word counts were analyzed across three OCR arrangements.


\begin{table*}[ht]
\centering
\caption{Word segmentation statistics under different OCR arrangements.}
\label{word_seg_table}
\scalebox{0.9}{
\begin{tabular}{l c c c c c}
\toprule
\textbf{Arrangement Method} & \textbf{Min} & \textbf{Max} & \textbf{Avg} & \textbf{Avg EM (\%)} & \textbf{Avg F1-score (\%)} \\ 
\midrule
Original Order                     & 0 & 11 & 0.56          & 18.45 & 40.62 \\
OCR Confidence        & 0 & 10 & 0.56          & 17.30 & 39.04 \\
TL-BR & 0 & 14 & \textbf{1.06} & \textbf{19.83} & \textbf{42.81} \\
\bottomrule
\end{tabular}
}
\end{table*}

As shown in Table \ref{word_seg_table}, the  top-left to bottom-
right (TL-BR) strategy yields the highest average number of segmented words and achieves the best model performance. This ordering aligns with natural Vietnamese reading patterns, allowing better context reconstruction and more accurate segmentation.

These results suggest that optimizing OCR arrangements to increase word segmentation quality can be an effective strategy for enhancing performance in Vietnamese VQA and other NLP tasks where word-level understanding is critical.

\section{Limitations and Future Work}
\label{sec8}
While the ViTextVQA dataset provides a resource for Vietnamese multimodal research, several limitations remain. First, the performance of current models is highly sensitive to OCR quality, which introduces errors that propagate into downstream predictions. Second, ViTextVQA primarily focuses on scene text understanding, which may limit its applicability in scenarios where visual reasoning extends beyond textual elements.

Additionally, the ViTextBLIP-2 model was trained specifically for Vietnamese and has not yet been evaluated in multilingual or zero-shot settings. Exploring cross-lingual transfer, robust OCR techniques, and visual reasoning beyond text-centric questions represents valuable directions for future research.

Looking forward, notable research directions can be highlighted, including:
\begin{itemize}
    \item Adapt the dataset for Visual Question Generation (VQG), where the task involves generating contextually relevant questions given an image and an answer \cite{zhang2017automatic, fan2018reinforcement, xu2020radial}. Beyond conventional VQA benchmarks, future work can extend these models to domains such as environmental and climate-related applications, which require advanced reasoning over multimodal information \cite{guo2023prediction, guo2024monthly, guo2024performance}. Considering recent findings on systematic poisoning attacks in healthcare ML systems \cite{mozaffari2014systematic}, security-conscious data adaptation will also be essential to ensure robustness against adversarial inputs.
    
    \item Investigate prompting-based methods with large language models (LLMs), inspired by promising results from recent studies \cite{jia2022visual, zhou2022learning}, which demonstrate strong performance even under limited fine-tuning. A natural extension is the development of a Vietnamese multimodal chatbot — similar in spirit to Flamingo \cite{alayrac2022flamingo}, GPT-4 \cite{achiam2023gpt}, and Gemini \cite{team2023gemini} — capable of handling image-grounded queries in Vietnamese, thereby advancing localized multimodal understanding and interaction.
\end{itemize}

\section{Conclusion}
\label{sec9}
In this work, we introduced ViTextVQA, a large-scale Vietnamese Visual Question Answering dataset, comprising over 16,000 images and 50,000 question-answer pairs. The dataset is designed to advance research in Vietnamese multimodal understanding, addressing the underrepresentation of low-resource languages in VQA. Unlike ViVQA, which relied on semi-automatic question generation, OpenViVQA, which only partially incorporated scene text, or ViOCRVQA, which omitted it entirely. We hope ViTextVQA will contribute meaningfully to the diversity of available linguistic resources and foster further research in Vietnamese language technologies.

We also benchmarked several state-of-the-art VQA models on ViTextVQA and observed that their performance, while strong on English datasets, does not translate equivalently to Vietnamese. This gap motivated a thorough performance analysis, uncovering key insights-most notably, that arranging OCR text in a top-left to bottom-right order significantly enhances model accuracy. Compared with baseline models such as BLIP-2, LaTr, and PreSTU, the proposed method ViTextBLIP-2 achieves substantial improvements, outperforming them by a margin of 53.95\% in overall F1-score on ViTextVQA. These findings highlight the need to tailor preprocessing and modeling strategies to the linguistic and structural characteristics of Vietnamese.

\section*{Acknowledgement}
This research is funded by Vietnam National University HoChiMinh City (VNU-HCM) under grant number NCM2025-26-02.

\section*{Author Contributions Statement}
 \textbf{Quan Van Nguyen:} Conceptualization; Formal analysis; Investigation; Methodology; Validation; Visualization; Writing - original draft.

 \textbf{Dan Quang Tran:} Conceptualization; Data curation; Formal analysis; Investigation; Validation; Visualization; Writing - original draft.
 
 \textbf{Huy Quang Pham:} Conceptualization; Data curation; Investigation; Methodology; Writing - original draft.
 
 \textbf{Thang Kien-Bao Nguyen:} Conceptualization; Data curation; Investigation; Methodology; Writing - original draft.
 
 \textbf{Nghia Hieu Nguyen:} Conceptualization; Data curation; Investigation; Methodology; Writing - review \& editing.
 
 \textbf{Kiet Van Nguyen:} Conceptualization; Formal analysis; Investigation; Methodology; Validation; Supervision; Writing - review \& editing.
 
 \textbf{Ngan Luu-Thuy Nguyen:} Conceptualization; Formal analysis; Investigation; Methodology; Validation; Supervision; Writing - review \& editing.

\section*{Declaration of Competing Interest}
The authors declare that they have no conflict of interest.

\section*{Data Availability}

Data will be made available on request.

\appendix

\section{Data creation rules}
\subsection*{General Rules}
General Rules are established to ensure that the process of creating questions and answers for images is done carefully and systematically. This regulation helps ensure consistency and quality of data, from ensuring the number of questions and answers per image.
\begin{itemize}
  \item \textbf{Rule 1:} For each subset, the annotator created at least 600 pairs of questions and answers.
  \item \textbf{Rule 2:} The annotator is encouraged to use their personal vocabulary to ask questions.
  \item \textbf{Rule 3:} Each image contains at most 10 pairs of questions and answers.
  \item \textbf{Rule 4:} All letters are in lowercase and in the scope of the standard keyboard.
\end{itemize}

\subsection*{Rules for creating questions}
The rules for creating questions provide specific instructions on how to ask questions so that they are specific, not vague, and ensure that the question leads to an answer that is already in the image. This way, not only does it create valuable questions, but it also creates convenience for participants to think and reason creatively and logically.
\begin{itemize}
  \item \textbf{Rule 1:} Question must be answered by text that appears only in the image.
  \item \textbf{Rule 2:} Do not ask questions with mathematical implications.
  \item \textbf{Rule 3:} Do not ask where the text is located.
  \item \textbf{Rule 4:} Do not ask options (yes/no).
  \item \textbf{Rule 5:} Do not ask the color of the text.
  \item \textbf{Rule 6:} Do not ask the number of text that the annotator has to count.
  \item \textbf{Rule 7:} Avoid asking questions that are too short.
  \item \textbf{Rule 8:} Avoid asking multiple questions that have the same answer in the same image.
  \item \textbf{Rule 9:} Avoid asking general questions that lead to many different answers.
\item \textbf{Rule 10:} Take the direction of the annotator seeing the image as the reference.
\end{itemize}

\subsection*{Rules for creating answers}
The rule for creating answers guides annotators to make sure that annotated answers are correct and do not contain extraneous information. This ensures the consistency of the dataset and makes the models easier to process.
\begin{itemize}
  \item \textbf{Rule 1:} The answer is only the words that appear in the image.
  \item \textbf{Rule 2:} Keep the content intact, convert everything to lowercase.
  \item \textbf{Rule 3:} If the text has a line, replace it with a space.
  \item \textbf{Rule 4:} If the answer has some words that are obscured, only annotate what the annotator sees.
\end{itemize}
\section{OCR System}

\begin{figure}[htp]
    \centering
    \includegraphics[width=1\textwidth]{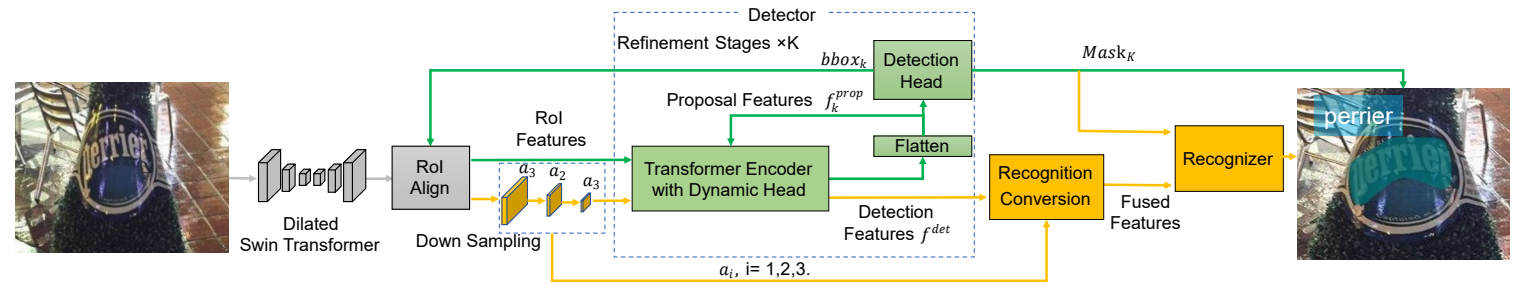}
    \caption{The overview of SwinTextSpotter structure}
    \label{ocr_system}
\end{figure}

\textbf{SwinTextSpotter: }
Since OCR quality strongly influences VQA performance, ViTextBLIP-2 incorporates SwinTextSpotter \cite{huang2022swintextspotter} as the OCR backbone due to its state-of-the-art accuracy and support for multilingual scene text spotting, including Vietnamese. SwinTextSpotter is an end-to-end framework that tightly integrates text detection and recognition using a shared Swin Transformer backbone, enhancing the synergy between these two stages (see Figure~\ref{ocr_system}).

In the present pipeline, the publicly available pre-trained SwinTextSpotter model is employed without additional fine-tuning. To enable language-specific customization, the text recognition decoder is configured with Vietnamese character sets and diacritic-aware settings. This adaptation ensures accurate transcription of tone-sensitive Vietnamese scene text, which is critical for downstream VQA tasks that require fine-grained linguistic understanding.

\bibliography{main}

@article{alayrac2022flamingo,
  title={Flamingo: a visual language model for few-shot learning},
  author={Alayrac, Jean-Baptiste and Donahue, Jeff and Luc, Pauline and Miech, Antoine and Barr, Iain and Hasson, Yana and Lenc, Karel and Mensch, Arthur and Millican, Katherine and Reynolds, Malcolm and others},
  journal={Advances in Neural Information Processing Systems},
  volume={35},
  pages={23716--23736},
  year={2022}
}

@article{rafiepour2023ctran,
  title={CTRAN: CNN-transformer-based network for natural language understanding},
  author={Rafiepour, Mehrdad and Sartakhti, Javad Salimi},
  journal={Engineering Applications of Artificial Intelligence},
  volume={126},
  pages={107013},
  year={2023},
  publisher={Elsevier}
}

@article{hu2024matryoshka,
  title={Matryoshka query transformer for large vision-language models},
  author={Hu, Wenbo and Dou, Zi-Yi and Li, Liunian and Kamath, Amita and Peng, Nanyun and Chang, Kai-Wei},
  journal={Advances in Neural Information Processing Systems},
  volume={37},
  pages={50168--50188},
  year={2024}
}

@inproceedings{wang2016cnn,
  title={Cnn-rnn: A unified framework for multi-label image classification},
  author={Wang, Jiang and Yang, Yi and Mao, Junhua and Huang, Zhiheng and Huang, Chang and Xu, Wei},
  booktitle={Proceedings of the IEEE conference on computer vision and pattern recognition},
  pages={2285--2294},
  year={2016}
}

@article{achiam2023gpt,
  title={Gpt-4 technical report},
  author={Achiam, Josh and Adler, Steven and Agarwal, Sandhini and Ahmad, Lama and Akkaya, Ilge and Aleman, Florencia Leoni and Almeida, Diogo and Altenschmidt, Janko and Altman, Sam and Anadkat, Shyamal and others},
  journal={arXiv preprint arXiv:2303.08774},
  year={2023}
}

@article{team2023gemini,
  title={Gemini: a family of highly capable multimodal models},
  author={Team, Gemini and Anil, Rohan and Borgeaud, Sebastian and Wu, Yonghui and Alayrac, Jean-Baptiste and Yu, Jiahui and Soricut, Radu and Schalkwyk, Johan and Dai, Andrew M and Hauth, Anja and others},
  journal={arXiv preprint arXiv:2312.11805},
  year={2023}
}

@inproceedings{tran2021vivqa,
  title={ViVQA: Vietnamese visual question answering},
  author={Tran, Khanh Quoc and Nguyen, An Trong and Le, An Tran-Hoai and Nguyen, Kiet},
  booktitle={Proceedings of the 35th Pacific Asia Conference on Language, Information and Computation},
  pages={683--691},
  year={2021}
}

@article{geigle2023mblip,
  title={mBLIP: Efficient Bootstrapping of Multilingual Vision-LLMs},
  author={Geigle, Gregor and Jain, Abhay and Timofte, Radu and Glava{\v{s}}, Goran},
  journal={arXiv preprint arXiv:2307.06930},
  year={2023}
}

@inproceedings{kil2023prestu,
  title={Prestu: Pre-training for scene-text understanding},
  author={Kil, Jihyung and Changpinyo, Soravit and Chen, Xi and Hu, Hexiang and Goodman, Sebastian and Chao, Wei-Lun and Soricut, Radu},
  booktitle={Proceedings of the IEEE/CVF International Conference on Computer Vision},
  pages={15270--15280},
  year={2023}
}

@inproceedings{singh2019towards,
  title={Towards vqa models that can read},
  author={Singh, Amanpreet and Natarajan, Vivek and Shah, Meet and Jiang, Yu and Chen, Xinlei and Batra, Dhruv and Parikh, Devi and Rohrbach, Marcus},
  booktitle={Proceedings of the IEEE/CVF conference on computer vision and pattern recognition},
  pages={8317--8326},
  year={2019}
}

@inproceedings{hu2020iterative,
  title={Iterative answer prediction with pointer-augmented multimodal transformers for textvqa},
  author={Hu, Ronghang and Singh, Amanpreet and Darrell, Trevor and Rohrbach, Marcus},
  booktitle={Proceedings of the IEEE/CVF conference on computer vision and pattern recognition},
  pages={9992--10002},
  year={2020}
}

@inproceedings{biten2022latr,
  title={Latr: Layout-aware transformer for scene-text vqa},
  author={Biten, Ali Furkan and Litman, Ron and Xie, Yusheng and Appalaraju, Srikar and Manmatha, R},
  booktitle={Proceedings of the IEEE/CVF conference on computer vision and pattern recognition},
  pages={16548--16558},
  year={2022}
}

@inproceedings{fang2023separate,
  title={Separate and locate: Rethink the text in text-based visual question answering},
  author={Fang, Chengyang and Li, Jiangnan and Li, Liang and Ma, Can and Hu, Dayong},
  booktitle={Proceedings of the 31st ACM International Conference on Multimedia},
  pages={4378--4388},
  year={2023}
}

@inproceedings{dosovitskiy2020image,
  title={An Image is Worth 16x16 Words: Transformers for Image Recognition at Scale},
  author={Dosovitskiy, Alexey and Beyer, Lucas and Kolesnikov, Alexander and Weissenborn, Dirk and Zhai, Xiaohua and Unterthiner, Thomas and Dehghani, Mostafa and Minderer, Matthias and Heigold, Georg and Gelly, Sylvain and others},
  booktitle={International Conference on Learning Representations},
  year={2021}
}

@inproceedings{biten2019scene,
  title={Scene text visual question answering},
  author={Biten, Ali Furkan and Tito, Ruben and Mafla, Andres and Gomez, Lluis and Rusinol, Mar{\c{c}}al and Valveny, Ernest and Jawahar, CV and Karatzas, Dimosthenis},
  booktitle={Proceedings of the IEEE/CVF international conference on computer vision},
  pages={4291--4301},
  year={2019}
}

@article{malinowski2014multi,
  title={A multi-world approach to question answering about real-world scenes based on uncertain input},
  author={Malinowski, Mateusz and Fritz, Mario},
  journal={Advances in neural information processing systems},
  volume={27},
  year={2014}
}

@inproceedings{antol2015vqa,
  title={Vqa: Visual question answering},
  author={Antol, Stanislaw and Agrawal, Aishwarya and Lu, Jiasen and Mitchell, Margaret and Batra, Dhruv and Zitnick, C Lawrence and Parikh, Devi},
  booktitle={Proceedings of the IEEE international conference on computer vision},
  pages={2425--2433},
  year={2015}
}

@inproceedings{goyal2017making,
  title={Making the v in vqa matter: Elevating the role of image understanding in visual question answering},
  author={Goyal, Yash and Khot, Tejas and Summers-Stay, Douglas and Batra, Dhruv and Parikh, Devi},
  booktitle={Proceedings of the IEEE conference on computer vision and pattern recognition},
  pages={6904--6913},
  year={2017}
}

@inproceedings{mishra2019ocr,
  title={Ocr-vqa: Visual question answering by reading text in images},
  author={Mishra, Anand and Shekhar, Shashank and Singh, Ajeet Kumar and Chakraborty, Anirban},
  booktitle={2019 international conference on document analysis and recognition (ICDAR)},
  pages={947--952},
  year={2019},
  organization={IEEE}
}

@inproceedings{mathew2021docvqa,
  title={Docvqa: A dataset for vqa on document images},
  author={Mathew, Minesh and Karatzas, Dimosthenis and Jawahar, CV},
  booktitle={Proceedings of the IEEE/CVF winter conference on applications of computer vision},
  pages={2200--2209},
  year={2021}
}

@inproceedings{tanaka2021visualmrc,
  title={Visualmrc: Machine reading comprehension on document images},
  author={Tanaka, Ryota and Nishida, Kyosuke and Yoshida, Sen},
  booktitle={Proceedings of the AAAI Conference on Artificial Intelligence},
  volume={35},
  number={15},
  pages={13878--13888},
  year={2021}
}

@article{nguyen2023vlsp,
  title={EVJVQA CHALLENGE: MULTILINGUAL VISUAL QUESTION ANSWERING},
  author={Nguyen, Ngan Luu-Thuy and Nguyen, Nghia Hieu and Vo, Duong TD and Tran, Khanh Quoc and Nguyen, Kiet},
  journal={Journal of Computer Science and Cybernetics},
  pages={237--258},
  year={2023}
}

@article{simonyan2014very,
  title={Very deep convolutional networks for large-scale image recognition},
  author={Simonyan, Karen and Zisserman, Andrew},
  journal={3rd International Conference on Learning Representations (ICLR 2015)},
  pages= {1--14},
  year={2015}
}

@article{hochreiter1997long,
  title={Long short-term memory},
  author={Hochreiter, Sepp and Schmidhuber, J{\"u}rgen},
  journal={Neural computation},
  volume={9},
  number={8},
  pages={1735--1780},
  year={1997},
  publisher={MIT press}
}

@inproceedings{szegedy2015going,
  title={Going deeper with convolutions},
  author={Szegedy, Christian and Liu, Wei and Jia, Yangqing and Sermanet, Pierre and Reed, Scott and Anguelov, Dragomir and Erhan, Dumitru and Vanhoucke, Vincent and Rabinovich, Andrew},
  booktitle={Proceedings of the IEEE conference on computer vision and pattern recognition},
  pages={1--9},
  year={2015}
}

@article{strobelt2017lstmvis,
  title={Lstmvis: A tool for visual analysis of hidden state dynamics in recurrent neural networks},
  author={Strobelt, Hendrik and Gehrmann, Sebastian and Pfister, Hanspeter and Rush, Alexander M},
  journal={IEEE transactions on visualization and computer graphics},
  volume={24},
  number={1},
  pages={667--676},
  year={2017},
  publisher={IEEE}
}

@inproceedings{xu2016ask,
  title={Ask, attend and answer: Exploring question-guided spatial attention for visual question answering},
  author={Xu, Huijuan and Saenko, Kate},
  booktitle = "Computer Vision - 14th European Conference, ECCV 2016, Proceedings",
  pages={451--466},
  year={2016},
  organization={Springer}
}

@article{zhang2010understanding,
  title={Understanding bag-of-words model: a statistical framework},
  author={Zhang, Yin and Jin, Rong and Zhou, Zhi-Hua},
  journal={International journal of machine learning and cybernetics},
  volume={1},
  pages={43--52},
  year={2010},
  publisher={Springer}
}

@inproceedings{he2016deep,
  title={Deep residual learning for image recognition},
  author={He, Kaiming and Zhang, Xiangyu and Ren, Shaoqing and Sun, Jian},
  booktitle={Proceedings of the IEEE conference on computer vision and pattern recognition},
  pages={770--778},
  year={2016}
}

@inproceedings{chung2014empirical,
title = "Empirical evaluation of gated recurrent neural networks on sequence modeling",
author = "Junyoung Chung and Caglar Gulcehre and Kyunghyun Cho and Yoshua Bengio",
year = "2014",
booktitle = "NIPS 2014 Workshop on Deep Learning, December 2014",
}

@inproceedings{anderson2018bottom,
  title={Bottom-up and top-down attention for image captioning and visual question answering},
  author={Anderson, Peter and He, Xiaodong and Buehler, Chris and Teney, Damien and Johnson, Mark and Gould, Stephen and Zhang, Lei},
  booktitle={Proceedings of the IEEE conference on computer vision and pattern recognition},
  pages={6077--6086},
  year={2018}
}

@inproceedings{devlin2018bert,
  title={Bert: Pre-training of deep bidirectional transformers for language understanding},
  author={Devlin, Jacob and Chang, Ming-Wei and Lee, Kenton and Toutanova, Kristina},
  booktitle={Proceedings of NAACL-HLT},
  pages={4171--4186},
  year={2019}
}

@article{lu2019vilbert,
  title={Vilbert: Pretraining task-agnostic visiolinguistic representations for vision-and-language tasks},
  author={Lu, Jiasen and Batra, Dhruv and Parikh, Devi and Lee, Stefan},
  journal={Advances in neural information processing systems},
  volume={32},
  year={2019}
}

@inproceedings{li2019visualbert,
  title={What does BERT with vision look at?},
  author={Li, Liunian Harold and Yatskar, Mark and Yin, Da and Hsieh, Cho-Jui and Chang, Kai-Wei},
  booktitle={Proceedings of the 58th Annual Meeting of the Association for Computational Linguistics},
  pages={5265--5275},
  year={2020}
}

@inproceedings{li2020unicoder,
  title={Unicoder-vl: A universal encoder for vision and language by cross-modal pre-training},
  author={Li, Gen and Duan, Nan and Fang, Yuejian and Gong, Ming and Jiang, Daxin},
  booktitle={Proceedings of the AAAI conference on artificial intelligence},
  volume={34},
  number={07},
  pages={11336--11344},
  year={2020}
}

@inproceedings{tan2019lxmert,
  title={Lxmert: Learning cross-modality encoder representations from transformers},
  author={Tan, Hao and Bansal, Mohit},
  booktitle={Proceedings of the 2019 Conference on Empirical Methods in Natural Language Processing and the 9th International Joint Conference on Natural Language Processing (EMNLP-IJCNLP)},
  pages={5100--5111},
  year={2019}
}

@inproceedings{su2019vl,
  title={VL-BERT: Pre-training of Generic Visual-Linguistic Representations},
  author={Su, Weijie and Zhu, Xizhou and Cao, Yue and Li, Bin and Lu, Lewei and Wei, Furu and Dai, Jifeng},
  booktitle={International Conference on Learning Representations},
  year={2019}
}

@inproceedings{chen2020uniter,
  title={Uniter: Universal image-text representation learning},
  author={Chen, Yen-Chun and Li, Linjie and Yu, Licheng and El Kholy, Ahmed and Ahmed, Faisal and Gan, Zhe and Cheng, Yu and Liu, Jingjing},
  booktitle={European conference on computer vision},
  pages={104--120},
  year={2020},
  organization={Springer}
}

@inproceedings{li2020oscar,
  title={Oscar: Object-semantics aligned pre-training for vision-language tasks},
  author={Li, Xiujun and Yin, Xi and Li, Chunyuan and Zhang, Pengchuan and Hu, Xiaowei and Zhang, Lei and Wang, Lijuan and Hu, Houdong and Dong, Li and Wei, Furu and others},
  booktitle={Computer Vision--ECCV 2020: 16th European Conference, Glasgow, UK, August 23--28, 2020, Proceedings, Part XXX 16},
  pages={121--137},
  year={2020},
  organization={Springer}
}

@inproceedings{girshick2014rich,
  title={Rich feature hierarchies for accurate object detection and semantic segmentation},
  author={Girshick, Ross and Donahue, Jeff and Darrell, Trevor and Malik, Jitendra},
  booktitle={Proceedings of the IEEE conference on computer vision and pattern recognition},
  pages={580--587},
  year={2014}
}

@inproceedings{li2022blip,
  title={Blip: Bootstrapping language-image pre-training for unified vision-language understanding and generation},
  author={Li, Junnan and Li, Dongxu and Xiong, Caiming and Hoi, Steven},
  booktitle={International Conference on Machine Learning},
  pages={12888--12900},
  year={2022},
  organization={PMLR}
}

@article{wang2022git,
  title={{GIT}: A Generative Image-to-text Transformer for Vision and Language},
  author={Wang, Jianfeng and Yang, Zhengyuan and Hu, Xiaowei and Li, Linjie and Lin, Kevin and Gan, Zhe and Liu, Zicheng and Liu, Ce and Wang, Lijuan},
  journal={Transactions on Machine Learning Research},
  year={2022}
}

@inproceedings{li-etal-2022-mplug,
    title = "m{PLUG}: Effective and Efficient Vision-Language Learning by Cross-modal Skip-connections",
    author = "Li, Chenliang  and
      Xu, Haiyang  and
      Tian, Junfeng  and
      Wang, Wei  and
      Yan, Ming  and
      Bi, Bin  and
      Ye, Jiabo  and
      Chen, He  and
      Xu, Guohai  and
      Cao, Zheng  and
      Zhang, Ji  and
      Huang, Songfang  and
      Huang, Fei  and
      Zhou, Jingren  and
      Si, Luo",
    editor = "Goldberg, Yoav  and
      Kozareva, Zornitsa  and
      Zhang, Yue",
    booktitle = "Proceedings of the 2022 Conference on Empirical Methods in Natural Language Processing",
    year = "2022",
    pages = "7241--7259"
}

@inproceedings{wang2023image,
  title={Image as a Foreign Language: BEiT Pretraining for Vision and Vision-Language Tasks},
  author={Wang, Wenhui and Bao, Hangbo and Dong, Li and Bjorck, Johan and Peng, Zhiliang and Liu, Qiang and Aggarwal, Kriti and Mohammed, Owais Khan and Singhal, Saksham and Som, Subhojit and others},
  booktitle={Proceedings of the IEEE/CVF Conference on Computer Vision and Pattern Recognition},
  pages={19175--19186},
  year={2023}
}

@inproceedings{chen2022pali,
  title={PaLI: Scaling Language-Image Learning in 100+ Languages},
  author={Chen, Xi and Wang, Xiao},
  booktitle={Conference on Neural Information Processing Systems (NeurIPS)},
  year={2022}
}

@inproceedings{phan2022vit5,
  title={ViT5: Pretrained Text-to-Text Transformer for Vietnamese Language Generation},
  author={Phan, Long and Tran, Hieu and Nguyen, Hieu and Trinh, Trieu H},
  booktitle={Proceedings of the 2022 Conference of the North American Chapter of the Association for Computational Linguistics: Human Language Technologies: Student Research Workshop},
  pages={136--142},
  year={2022}
}

@inproceedings{zhang2021vinvl,
  title={Vinvl: Revisiting visual representations in vision-language models},
  author={Zhang, Pengchuan and Li, Xiujun and Hu, Xiaowei and Yang, Jianwei and Zhang, Lei and Wang, Lijuan and Choi, Yejin and Gao, Jianfeng},
  booktitle={Proceedings of the IEEE/CVF conference on computer vision and pattern recognition},
  pages={5579--5588},
  year={2021}
}

@inproceedings{huang2022swintextspotter,
  title={Swintextspotter: Scene text spotting via better synergy between text detection and text recognition},
  author={Huang, Mingxin and Liu, Yuliang and Peng, Zhenghao and Liu, Chongyu and Lin, Dahua and Zhu, Shenggao and Yuan, Nicholas and Ding, Kai and Jin, Lianwen},
  booktitle={proceedings of the IEEE/CVF conference on computer vision and pattern recognition},
  pages={4593--4603},
  year={2022}
}

@article{kingma2014adam,
  title={Adam: A method for stochastic optimization},
  author={Kingma, Diederik P and Ba, Jimmy},
  journal={International Conference on Learning Representations (ICLR)},
  year={2015}
}

@inproceedings{wang2022chiqa,
  title={Chiqa: A large scale image-based real-world question answering dataset for multi-modal understanding},
  author={Wang, Bingning and Lv, Feiyang and Yao, Ting and Ma, Jin and Luo, Yu and Liang, Haijin},
  booktitle={Proceedings of the 31st ACM International Conference on Information \& Knowledge Management},
  pages={1996--2006},
  year={2022}
}

@inproceedings{qi2022dureadervis,
  title={Dureadervis: A: A chinese dataset for open-domain document visual question answering},
  author={Qi, Le and Lv, Shangwen and Li, Hongyu and Liu, Jing and Zhang, Yu and She, Qiaoqiao and Wu, Hua and Wang, Haifeng and Liu, Ting},
  booktitle={Findings of the Association for Computational Linguistics: ACL 2022},
  pages={1338--1351},
  year={2022}
}

@article{tran2023viclevr,
  title={Viclevr: A visual reasoning dataset and hybrid multimodal fusion model for visual question answering in vietnamese},
  author={Tran, Khiem Vinh and Phan, Hao Phu and Nguyen, Kiet and Nguyen, Ngan Luu Thuy},
  journal={Multimedia Systems},
  volume={30},
  number={4},
  pages={199},
  year={2024},
  publisher={Springer}
}

@article{nguyen2023openvivqa,
  title={Openvivqa: Task, dataset, and multimodal fusion models for visual question answering in vietnamese},
  author={Nguyen, Nghia Hieu and Vo, Duong TD and Van Nguyen, Kiet and Nguyen, Ngan Luu-Thuy},
  journal={Information Fusion},
  volume={100},
  pages={101868},
  year={2023},
  publisher={Elsevier}
}

@inproceedings{li2023blip,
  title={Blip-2: Bootstrapping language-image pre-training with frozen image encoders and large language models},
  author={Li, Junnan and Li, Dongxu and Savarese, Silvio and Hoi, Steven},
  booktitle={International conference on machine learning},
  pages={19730--19742},
  year={2023},
  organization={PMLR}
}

@inproceedings{vu2018vncorenlp,
  title={VnCoreNLP: A Vietnamese Natural Language Processing Toolkit},
  author={Vu, Thanh and Nguyen, Dat Quoc and Dras, Mark and Johnson, Mark and others},
  booktitle={Proceedings of the 2018 Conference of the North American Chapter of the Association for Computational Linguistics: Demonstrations},
  pages={56--60},
  year={2018}
}

@inproceedings{pires2019multilingual,
  title={How Multilingual is Multilingual BERT?},
  author={Pires, Telmo and Schlinger, Eva and Garrette, Dan},
  booktitle={Proceedings of the 57th Annual Meeting of the Association for Computational Linguistics},
  pages={4996--5001},
  year={2019}
}

@inproceedings{jia2022visual,
  title={Visual prompt tuning},
  author={Jia, Menglin and Tang, Luming and Chen, Bor-Chun and Cardie, Claire and Belongie, Serge and Hariharan, Bharath and Lim, Ser-Nam},
  booktitle={European Conference on Computer Vision},
  pages={709--727},
  year={2022},
  organization={Springer}
}

@article{zhou2022learning,
  title={Learning to prompt for vision-language models},
  author={Zhou, Kaiyang and Yang, Jingkang and Loy, Chen Change and Liu, Ziwei},
  journal={International Journal of Computer Vision},
  volume={130},
  number={9},
  pages={2337--2348},
  year={2022},
  publisher={Springer}
}

@inproceedings{rajpurkar-etal-2016-squad,
    title = "{SQ}u{AD}: 100,000+ Questions for Machine Comprehension of Text",
    author = "Rajpurkar, Pranav  and
      Zhang, Jian  and
      Lopyrev, Konstantin  and
      Liang, Percy",
    editor = "Su, Jian  and
      Duh, Kevin  and
      Carreras, Xavier",
    booktitle = "Proceedings of the 2016 Conference on Empirical Methods in Natural Language Processing",
    month = nov,
    year = "2016",
    address = "Austin, Texas",
    publisher = "Association for Computational Linguistics",
    pages = "2383--2392",
}

@article{Qwen-VL,
  title={Qwen-VL: A Versatile Vision-Language Model for Understanding, Localization, Text Reading, and Beyond},
  author={Bai, Jinze and Bai, Shuai and Yang, Shusheng and Wang, Shijie and Tan, Sinan and Wang, Peng and Lin, Junyang and Zhou, Chang and Zhou, Jingren},
  journal={arXiv preprint arXiv:2308.12966},
  year={2023}
}

@article{Nguyen_Nguyen_Ngo_Vu_Tran_Ngo_Le_2019, 
title={VLSP SHARED TASK: SENTIMENT ANALYSIS}, 
volume={34}, 
number={4}, 
journal={Journal of Computer Science and Cybernetics}, 
author={Nguyen, Huyen T M and Nguyen, Hung V and Ngo, Quyen T and Vu, Luong X and Tran, Vu Mai and Ngo, Bach X and Le, Cuong A}, 
year={2019}, 
month={Jan.}, 
pages={295–310} }

@inproceedings{nguyen-etal-2018-fast,
    title = "A Fast and Accurate {V}ietnamese Word Segmenter",
    author = "Nguyen, Dat Quoc  and
      Nguyen, Dai Quoc  and
      Vu, Thanh  and
      Dras, Mark  and
      Johnson, Mark",
    editor = "Calzolari, Nicoletta  and
      Choukri, Khalid  and
      Cieri, Christopher  and
      Declerck, Thierry  and
      Goggi, Sara  and
      Hasida, Koiti  and
      Isahara, Hitoshi  and
      Maegaard, Bente  and
      Mariani, Joseph  and
      Mazo, H{\'e}l{\`e}ne  and
      Moreno, Asuncion  and
      Odijk, Jan  and
      Piperidis, Stelios  and
      Tokunaga, Takenobu",
    booktitle = "Proceedings of the Eleventh International Conference on Language Resources and Evaluation ({LREC} 2018)",
    month = may,
    year = "2018",
    address = "Miyazaki, Japan",
    publisher = "European Language Resources Association (ELRA)",
}

@inproceedings{zhang2017automatic,
  title={Automatic generation of grounded visual questions},
  author={Zhang, Shijie and Qu, Lizhen and You, Shaodi and Yang, Zhenglu and Zhang, Jiawan},
  booktitle={International Joint Conference on Artificial Intelligence},
  pages={4235--4243},
  year={2017}
}

@inproceedings{fan2018reinforcement,
  title={A reinforcement learning framework for natural question generation using bi-discriminators},
  author={Fan, Zhihao and Wei, Zhongyu and Wang, Siyuan and Liu, Yang and Huang, Xuan-Jing},
  booktitle={Proceedings of the 27th International Conference on Computational Linguistics},
  pages={1763--1774},
  year={2018}
}

@article{xu2020radial,
  title={Radial graph convolutional network for visual question generation},
  author={Xu, Xing and Wang, Tan and Yang, Yang and Hanjalic, Alan and Shen, Heng Tao},
  journal={IEEE transactions on neural networks and learning systems},
  volume={32},
  number={4},
  pages={1654--1667},
  year={2020},
  publisher={IEEE}
}

@article{pham2025viocrvqa,
  title={ViOCRVQA: novel benchmark dataset and VisionReader for visual question answering by understanding Vietnamese text in images},
  author={Pham, Huy Quang and Nguyen, Thang Kien-Bao and Van Nguyen, Quan and Tran, Dan Quang and Nguyen, Nghia Hieu and Van Nguyen, Kiet and Nguyen, Ngan Luu-Thuy},
  journal={Multimedia Systems},
  volume={31},
  number={2},
  pages={106},
  year={2025},
  publisher={Springer}
}

@article{guo2023prediction,
  title={Prediction of monthly average and extreme atmospheric temperatures in Zhengzhou based on artificial neural network and deep learning models},
  author={Guo, Qingchun and He, Zhenfang and Wang, Zhaosheng},
  journal={Frontiers in Forests and Global Change},
  volume={6},
  pages={1249300},
  year={2023},
  publisher={Frontiers Media SA}
}

@article{guo2024monthly,
  title={Monthly climate prediction using deep convolutional neural network and long short-term memory},
  author={Guo, Qingchun and He, Zhenfang and Wang, Zhaosheng},
  journal={Scientific Reports},
  volume={14},
  number={1},
  pages={17748},
  year={2024},
  publisher={Nature Publishing Group UK London}
}

@article{guo2024performance,
  title={A Performance Comparison Study on Climate Prediction in Weifang City Using Different Deep Learning Models},
  author={Guo, Qingchun and He, Zhenfang and Wang, Zhaosheng and Qiao, Shuaisen and Zhu, Jingshu and Chen, Jiaxin},
  journal={Water},
  volume={16},
  number={19},
  pages={2870},
  year={2024},
  publisher={MDPI}
}

@article{gao2015you,
  title={Are you talking to a machine? dataset and methods for multilingual image question},
  author={Gao, Haoyuan and Mao, Junhua and Zhou, Jie and Huang, Zhiheng and Wang, Lei and Xu, Wei},
  journal={Advances in neural information processing systems},
  volume={28},
  year={2015}
}

@article{kamel2023vaqa,
  title={Vaqa: Visual arabic question answering},
  author={Kamel, Sarah M and Hassan, Shimaa I and Elrefaei, Lamiaa},
  journal={Arabian Journal for Science and engineering},
  volume={48},
  number={8},
  pages={10803--10823},
  year={2023},
  publisher={Springer}
}

@ARTICLE{10878995,
  author={Bhuyan, Md. Shalha Mucha and Hossain, Eftekhar and Sathi, Khaleda Akhter and Hossain, Md. Azad and Dewan, M. Ali Akber},
  journal={IEEE Access}, 
  title={BVQA: Connecting Language and Vision Through Multimodal Attention for Open-Ended Question Answering}, 
  year={2025},
  volume={13},
  number={},
  pages={27570-27586},
  doi={10.1109/ACCESS.2025.3540388}}

@article{mozaffari2014systematic,
  title={Systematic poisoning attacks on and defenses for machine learning in healthcare},
  author={Mozaffari-Kermani, Mehran and Sur-Kolay, Susmita and Raghunathan, Anand and Jha, Niraj K},
  journal={IEEE journal of biomedical and health informatics},
  volume={19},
  number={6},
  pages={1893--1905},
  year={2014},
  publisher={IEEE}
}

\bibliographystyle{elsarticle-num-names}







\end{document}